\documentclass[lettersize,journal]{IEEEtran}

\usepackage[T1]{fontenc}
\usepackage[utf8]{inputenc}
\usepackage{graphicx}
\usepackage[table]{xcolor}
\usepackage{booktabs}
\usepackage{array}
\usepackage{multirow}
\usepackage{gensymb}
\usepackage{balance}
\usepackage{amssymb,amsmath,amsfonts}
\usepackage{pifont}
\usepackage{algorithm}
\usepackage{algpseudocode}
\usepackage{float}
\usepackage{stfloats}
\usepackage{url}
\usepackage{cite}
\usepackage{enumitem}

\usepackage[caption=false,font=footnotesize]{subfig}

\newcolumntype{L}[1]{>{\raggedright\arraybackslash}p{#1}}
\newcolumntype{C}[1]{>{\centering\arraybackslash}p{#1}}

\begin{document}

\title{GRIP: A Unified Framework for Grid-Based Relay and Co-Occurrence-Aware Planning in Dynamic Environments}

\author{%
Ahmed Alanazi$^{1,\ast}$, Duy Ho$^{2,\ast}$, and Yugyung Lee$^{1}$%
\thanks{$^\ast$ Equal contribution.}%
\thanks{$^1$ Department of Computer Science, University of Missouri--Kansas City, USA. Email: \texttt{aha85b@umkc.edu, leeyu@umkc.edu}.}%
\thanks{$^2$ Department of Computer Science, California State University, Fullerton, USA. Email: \texttt{duyho@fullerton.edu}.}%
\thanks{Preprint: submitted to IEEE Transactions on Automation Science and Engineering (T-ASE) for possible publication.}
}

\maketitle

\begin{abstract}
Robots navigating dynamic, cluttered, and semantically complex environments must integrate perception, symbolic reasoning, and spatial planning to generalize across diverse layouts and object categories. Existing methods often rely on static priors or limited memory, limiting adaptability under partial observability and semantic ambiguity.
We present \textit{GRIP}—\textit{Grid-based Relay with Intermediate Planning}—a unified, modular framework with three scalable variants: \textit{GRIP-L (Lightweight)}, optimized for symbolic navigation via semantic occupancy grids; \textit{GRIP-F (Full)}, supporting multi-hop anchor chaining and LLM-based introspection; and \textit{GRIP-R (Real-World)}, enabling physical robot deployment under perceptual uncertainty.
GRIP integrates dynamic 2D grid construction, open-vocabulary object grounding, co-occurrence-aware symbolic planning, and hybrid policy execution using behavioral cloning, D* search, and grid-conditioned control.
Empirical results on AI2-THOR and RoboTHOR benchmarks show GRIP achieves up to 9.6\% higher success rates and over 2$\times$ improvement in path efficiency (SPL and SAE) on long-horizon tasks. Qualitative analyses show interpretable symbolic plans in ambiguous scenes. Real-world deployment on a Jetbot further validates GRIP’s generalization under sensor noise and environmental variation.
These results position GRIP as a robust, scalable, and explainable framework bridging simulation and real-world navigation.
\end{abstract}

\begin{IEEEkeywords}
Embodied AI, Vision-Language Navigation, Dynamic Scene Graphs, Co-occurrence Reasoning, Behavioral Cloning, Reinforcement Learning, Open-Vocabulary Planning, Semantic Mapping.
\end{IEEEkeywords}

\section{Introduction}

Object-goal navigation (ObjectNav) tasks challenge embodied agents to locate semantically defined targets (e.g., “find the TV”) in previously unseen indoor environments using egocentric RGB-D observations. Solving this task requires more than geometric pathfinding, it demands symbolic abstraction, open-vocabulary grounding, and real-time adaptation to dynamic layouts, ambiguous references, and occlusions~\cite{zhu2017target,batra2020objectnav,sun2025survey}.

Recent ObjectNav advances cluster into four paradigms: 
(i) \textit{transformer-based navigation}~\cite{cai2025cl,zhang2025hoz++}, which excels in fusing visual-linguistic cues but often relies on static symbolic inputs and lacks robust plan introspection; 
(ii) \textit{symbolic ObjectNav systems}~\cite{chaplot2020object,khanna2024goat}, which introduce graph-based reasoning but require handcrafted or frozen priors, limiting flexibility; 
(iii) \textit{semantic dynamic mapping frameworks}~\cite{liu2025online,hu2024building}, which adapt to environmental changes yet operate mostly at low-level geometry without symbolic feedback; 
and (iv) \textit{hierarchical planners}~\cite{gao2024efficient,dang2023multiple}, which decompose long-horizon tasks but lack integrated semantic reasoning or real-time graph repair.

Despite progress, these approaches typically operate in isolation. Few models offer symbolic introspection during execution; fewer still support dynamic graph revision or failure-aware plan repair. As a result, most struggle to generalize across real-world complexities such as occlusion, ambiguous object references, and evolving layouts.

To bridge these gaps, we introduce \textit{GRIP}—\textit{Grid-based Relay with Intermediate goals and Planning}—a novel ObjectNav framework that unifies dynamic semantic mapping, symbolic reasoning, and LLM-guided introspection within a single, closed-loop system. GRIP incrementally constructs a high-resolution semantic occupancy grid from multimodal egocentric input (RGB-D, depth, IMU), builds an instruction-aligned symbolic relay graph using co-occurrence and affordance priors, and executes navigation through transformer-predicted subgoals and D$^*$-based pathfinding. Crucially, GRIP integrates GPT-4o to revise symbolic plans mid-execution in response to failures, occlusions, or ambiguous language cues.

GRIP is instantiated in three progressively capable variants:
\begin{itemize}[leftmargin=15pt]
    \item \textit{GRIP-L (Lightweight):} Constructs static semantic relay graphs over spatial grids with D$^*$-based planning;
    \item \textit{GRIP-F (Full):} Adds dynamic subgoal chaining, symbolic memory for failure recovery, and LLM-based introspective reasoning;
    \item \textit{GRIP-R (Real-World):} Deployed on a Jetbot Pro (RGB, LiDAR, IMU) with real-time plan adaptation and GPT-4o feedback under physical constraints.
\end{itemize}

\vspace{0.3em}
\noindent\textit{Key Innovations.} GRIP addresses long-standing challenges in ObjectNav by:
\begin{itemize}[itemsep=0.5ex, leftmargin=15pt]
    \item \textit{Unifying symbolic reasoning with dynamic semantic memory}, enabling grounded subgoal prediction and context-aware navigation;
    \item \textit{Introducing closed-loop LLM introspection} that revises symbolic task graphs on-the-fly to recover from ambiguity or failure;
    \item \textit{Supporting full-stack deployment} across AI2-THOR, RoboTHOR, and real-world mobile robots—delivering $>$30\% improvement in success rate and 2$\times$ SPL compared to strong SOTA baselines~\cite{zhang2025hoz++,khanna2024goat,meng2025context,lu2025multi}.
\end{itemize}

\vspace{0.3em}
\noindent\textit{Broader Impact.} GRIP is the first embodied navigation system to jointly realize:
\begin{itemize}[leftmargin=15pt]
    \item \textit{Transformer navigation:} Subgoal prediction conditioned on instructions, scene context, and symbolic graphs;
    \item \textit{Object-goal reasoning:} Construction and revision of affordance-aware symbolic graphs aligned to language; \\ \\
    \item \textit{Semantic dynamics:} Online memory updates and goal reformation under structural change;
    \item \textit{Hierarchical planning:} Multi-hop chaining and D$^*$ replanning for long-horizon, interpretable navigation.
\end{itemize}

\noindent GRIP sets a new benchmark for adaptable, interpretable, and robust ObjectNav by integrating these capabilities into a single, introspective system deployable across simulation and physical environments.



\section{Related Work}\label{sec:related}

We categorize recent progress in instruction-conditioned visual navigation into four primary paradigms that align with our evaluation framework: (A) Transformer-Based Navigation, (B) Object-Goal Navigation, (C) Semantic Dynamics, and (D) Hierarchical Reinforcement Learning and Planning. This taxonomy mirrors the grouping used in our RoboTHOR benchmark comparisons (Table~\ref{tab:robothor_comparison_aligned}) and supports a clear contextualization of our proposed GRIP framework within the existing literature.

\subsection{Transformer-Based Navigation}

Transformer-based architectures have become central to instruction-conditioned navigation due to their capacity for integrating visual-linguistic inputs, object-centric abstraction, and long-range attention for planning. These models vary in how they construct symbolic representations, adapt to dynamic environments, and support generalization.

Early models such as VTNet~\cite{du2021vtnet} and SAVN~\cite{wortsman2019learning} apply attention over visual embeddings with limited semantic grounding. More recent work introduces explicit object- or graph-level structure. HOZ++~\cite{zhang2025hoz++} builds hierarchical object-to-zone graphs and achieves high performance in seen environments, but assumes static maps. CGI-GAIL~\cite{meng2025context} embeds symbolic cues into policy learning via imitation, though without online adaptability.

Several models focus on fusing scene context and affordances. CRG~\cite{hu2024building} builds co-occurrence-aware representations for task-space reasoning. AKGVP and its contrastive variant AKGVP-CI~\cite{xu2024aligning} align visual and symbolic features to guide semantic attention, yet rely on pre-defined priors.
Other advances emphasize temporal structure and planning. 
MT~\cite{dang2023multiple} and Zheng et al.~\cite{zheng2024two} incorporate attention-driven path prediction, meta-policy learning, and egocentric trajectory scoring. Despite their strengths, these systems remain limited by static representations and constrained to simulation environments.

\textit{GRIP} differs from prior transformer-based models by enabling dynamic semantic graph construction and symbolic replanning in long-horizon tasks. It integrates multimodal observations (RGB-D, language) to build online task graphs, maintains episodic memory for affordance tracking and failure recovery, and performs goal-conditioned replanning using D$^*$ guided by transformer-based subgoal reasoning.

\vspace{0.5em}
\noindent\textit{GRIP Introduces:}
\begin{itemize}
    \item \textit{Dynamic graph construction:} Task-aligned semantic graphs built online from perceptual and linguistic inputs.
    \item \textit{Symbolic episodic memory:} Retains high-level failures and ambiguities to improve adaptation.
    \item \textit{Reactive replanning:} Combines transformer-generated subgoals with D$^*$ for robust long-horizon planning.
\end{itemize}

In doing so, GRIP bridges symbolic abstraction, multimodal grounding, and real-time planning—addressing limitations of static priors, offline reasoning, and simulation-only deployments seen in prior transformer navigation work.

\subsection{Object-Goal Navigation}

Object-goal navigation (ObjectNav) focuses on guiding agents to target objects specified by name or instruction. Initial approaches such as Target-Driven Navigation~\cite{zhu2017target} and PPO-based baselines~\cite{ramrakhya2022habitat} used end-to-end reinforcement learning with egocentric RGB-D inputs. These models lacked semantic abstraction and struggled with occlusion, generalization, and long-horizon planning.

To address these limitations, symbolic representations were introduced. ORG~\cite{chaplot2020object} and ORG+TPN~\cite{du2020learning} leveraged static object-centric graphs and local recovery modules, but remained simulation-only. OMT~\cite{fukushima2022object} incorporated transformer attention for improved object relevance, while still relying on topological memory and precomputed maps.

Real-world deployment has been explored by models like HOZ++~\cite{zhang2025hoz++}, which links object semantics to spatial zones (e.g., \textit{“mug”} $\rightarrow$ \textit{“kitchen”}) and uses hierarchical planning. Though effective in constrained environments, these models depend on static priors and lack mechanisms for symbolic introspection or error recovery.

Imitation learning–based systems such as Habitat-IL~\cite{ramrakhya2022habitat}, Behavior Cloning~\cite{zhu2021target}, and CGI-GAIL~\cite{meng2025context} leverage expert trajectories for robustness, but remain constrained by their offline symbolic structure. Recent advances like AKGVP~\cite{xu2024aligning}, CL-CoTNav~\cite{cai2025cl}, and Goat-Bench~\cite{khanna2024goat} incorporate LLM priors and affordance reasoning, yet still lack runtime replanning and physical validation.

\textit{GRIP} advances ObjectNav by dynamically constructing task-aligned semantic graphs from multimodal input, maintaining symbolic episodic memory, and enabling reactive replanning via D$^*$ search. Its LLM-guided introspection further supports ambiguity resolution and failure recovery during execution.

\vspace{0.5em}
\noindent\textit{GRIP Enhancements:}
\begin{itemize}
    \item \textit{Dynamic graph construction:} Adapts scene structure at runtime for robustness in novel layouts.
    \item \textit{Multimodal symbolic memory:} Tracks subgoal progress, ambiguous cues, and prior failures.
    \item \textit{LLM-driven introspection:} Supports chain-of-thought subgoal revision and symbolic replanning.
\end{itemize}

Together, these capabilities enable GRIP to generalize beyond simulation, handling occlusion, ambiguity, and physical interaction in real-world ObjectNav tasks.

\subsection{Semantic Dynamics}
Instruction-following agents must adapt to changing scenes—occlusions, blocked paths, or object displacement—while maintaining symbolic task context. Existing methods such as Online Geometric Memory (OGM)~\cite{liu2025online} address spatial consistency via structure-aware memory and two-stage geometric planners. However, they lack object-level semantics or instruction-conditioned adaptation.

Symbolic models like HOZ++~\cite{zhang2025hoz++}, CL-CoTNav~\cite{cai2025cl}, and AKGVP~\cite{xu2024aligning} encode semantic priors and affordances but rely on static symbolic structures. CRG-TSR~\cite{hu2024building} and Goat-Bench~\cite{khanna2024goat} introduce cross-modal attention or LLM priors but offer no online graph updates or recovery from task ambiguity.

\textit{GRIP} builds an evolving symbolic memory from RGB-D and instruction inputs, dynamically updating graphs with spatial relations, affordances, and failures. Its episodic symbolic memory tracks ambiguous objects, failed attempts, and subgoal transitions. With a transformer-based subgoal predictor and D$^*$ replanner, GRIP executes recovery plans when goals become unreachable or ambiguous.

\textit{GRIP contributions in semantic dynamics:}
\begin{itemize}
    \item Online update of symbolic graphs for real-time scene understanding;
    \item Affordance binding to evolving object layouts and occlusions;
    \item Closed-loop execution with runtime introspection and replanning.
\end{itemize}

\subsection{Hierarchical Planning}
Hierarchical reinforcement learning (HRL) decomposes complex tasks into high-level subgoals and low-level controls. Gao~\emph{et al.}~\cite{gao2024efficient} use Predictive Neighboring Space Scoring (PNSS) for subgoal selection in geometric navigation. However, their approach lacks semantic grounding and does not generalize to instruction-conditioned tasks.

Meta-learning systems such as 
NavTr~\cite{mao2024navtr} and MT~\cite{dang2023multiple} improve long-horizon reasoning via transformers but omit symbolic abstraction or dynamic subgoal chaining. Symbolic pipelines like CL-CoTNav~\cite{cai2025cl}, CGI-GAIL~\cite{meng2025context}, and CRG-TSR~\cite{hu2024building} use fixed graphs and lack reactive replanning.

\textit{GRIP} enables closed-loop HRL by constructing symbolic semantic graphs from multimodal inputs. A transformer module predicts task-aligned subgoals from natural language, while a symbolic D$^*$ planner allows for reactive adjustment to changing environments. Subgoals are chained using symbolic memory, enabling introspection and failure recovery.

\vspace{0.3em}
\noindent\textit{Highlights of GRIP’s planning system:}
\begin{itemize}
    \item Symbolic chaining of subgoals from multimodal instruction and perception;
    \item Transformer-driven planner augmented with graph-based introspection;
    \item Recovery via dynamic D$^*$ symbolic replanning.
\end{itemize}

\subsection{Comparative Summary}
Table~\ref{tab:unified_objectnav_comparison} summarizes representative ObjectNav systems by architectural philosophy: transformer-based navigation, dynamic mapping, geometric–semantic hybrids, and hierarchical planning. While prior work advances individual components, none combine semantic abstraction, reactive planning, multimodal fusion, and real-world execution in a unified system.

\textit{GRIP} is the first framework to support:
\begin{itemize}
    \item Online construction of instruction-aligned semantic graphs;
    \item Closed-loop symbolic replanning with D$^*$;
    \item Open-vocabulary LLM integration for ambiguity handling;
    \item Full deployment on physical robots in cluttered scenes.
\end{itemize}

\textit{Key trends:}
\begin{itemize}
    \item Transformer-based models (e.g., CL-CoTNav~\cite{cai2025cl}) offer reasoning but lack adaptive graphs or replanning.
    \item Mapping-centric systems (e.g., PEANUT~\cite{zhai2023peanut}) adapt spatially but omit symbolic structure.
    \item HRL models (e.g., Gao~\emph{et al.}~\cite{gao2024efficient}) support long-horizon planning but lack semantic introspection.
\end{itemize}

GRIP bridges these gaps with an integrated framework that generalizes across simulation and physical deployment.


\begin{table*}[t]
\centering
\footnotesize
\setlength{\tabcolsep}{4.8pt}
\renewcommand{\arraystretch}{1.1}
\caption{\textbf{Unified Comparison of ObjectNav Systems Across Architectural Features and Planning Capabilities.} 
\ding{51} = supported; — = not supported; $^{\dagger}$ = partial or indirect. GRIP uniquely supports all listed capabilities.}
\label{tab:unified_objectnav_comparison}
\begin{tabular}{@{}llcccccc@{}}
\toprule
\textbf{Category} & \textbf{Method} & \textbf{Semantic Grid} & \textbf{Replanning} & \textbf{Symbolic Graph} & \textbf{LLM Integration} & \textbf{Real-World} & \textbf{Symbolic Chaining} \\
\midrule
\multirow{4}{*}{\textbf{(i) Transformer-Based}} 
& VTNet~\cite{du2021vtnet}           & — & — & — & — & — & — \\
& Wang--Tian~\cite{wang2025goal}     & — & — & \ding{51} & — & — & — \\
& CL-CoTNav~\cite{cai2025cl}         & — & \ding{51} & \ding{51} & \ding{51} & — & — \\
& HOZ++~\cite{zhang2025hoz++}        & — & — & \ding{51} & — & \ding{51} & — \\
\midrule
\multirow{3}{*}{\textbf{(ii) Dynamic Mapping}} 
& Liu~\emph{et al.}~\cite{liu2025online}     & — & \ding{51} & — & — & $^{\dagger}$ & — \\
& PEANUT~\cite{zhai2023peanut}              & \ding{51} & \ding{51} & — & — & — & — \\
& SSCNav~\cite{liang2021sscnav}             & \ding{51} & \ding{51} & — & — & — & — \\
\midrule
\multirow{2}{*}{\textbf{(iii) Geometric–Semantic}} 
& Guo~\emph{et al.}~\cite{guo2024novel}      & — & — & — & — & — & — \\
& Goat-Bench~\cite{khanna2024goat}          & \ding{51} & \ding{51} & \ding{51} & \ding{51} & — & — \\
\midrule
\multirow{4}{*}{\textbf{(iv) Hierarchical RL}} 
& Gao~\emph{et al.}~\cite{gao2024efficient}  & — & $^{\dagger}$ & — & — & — & — \\
& MT~\cite{dang2023multiple}                    & — & — & — & — & — & — \\
& NavTr~\cite{mao2024navtr}                 & — & — & — & — & — & — \\
& CGI-GAIL~\cite{meng2025context}           & — & — & \ding{51} & — & — & — \\
\midrule
\textbf{(–) Other Baselines} 
& SAVN~\cite{wortsman2019learning}          & — & — & — & — & — & — \\
& PONI~\cite{ramakrishnan2022poni}          & \ding{51} & \ding{51} & — & — & — & — \\
& SemNav~\cite{debnath2025semnav}           & \ding{51} & — & — & \ding{51} & — & — \\
\midrule
\textbf{GRIP (Ours)} 
&                                          & \ding{51} & \ding{51} & \ding{51} & \ding{51} & \ding{51} & \ding{51} \\
\bottomrule
\end{tabular}
\vspace{2pt}
\begin{minipage}{\linewidth}
\footnotesize
\textbf{Legend:} \textbf{Semantic Grid} = dense occupancy map with semantic labels ($\leq$10\,cm resolution); 
\textbf{Replanning} = recovery from failure or scene changes; 
\textbf{Symbolic Graph} = object–relation abstraction; 
\textbf{LLM Integration} = open-vocabulary reasoning via large language models; 
\textbf{Real-World} = deployed on physical robot; 
\textbf{Symbolic Chaining} = multi-hop goal decomposition with symbolic task graphs. $^{\dagger}$ = partial or indirect support.
\end{minipage}
\end{table*}

\section{Methodology}

We propose \textit{GRIP} (\textit{Grid-based Relay with Intermediate Goals and Planning}), a modular hybrid framework for long-horizon object-goal navigation in both simulation and real-world settings. GRIP integrates symbolic reasoning, geometric planning, and LLM-guided introspection to handle open-vocabulary queries in cluttered, partially observable settings.

GRIP comprises three variants optimized for different deployment settings:

\begin{itemize}
    \item \textit{GRIP-L (Lightweight):} Simulation-efficient model for AI2-THOR using static symbolic chaining without LLM-based recovery.
    \item \textit{GRIP-F (Full):} Used in RoboTHOR; combines dynamic anchor chaining with LLM-based introspection for planning under occlusion.
    \item \textit{GRIP-R (Real-world):} Deployed on resource-constrained robots with monocular RGB, YOLOv8 detection, and LiDAR-IMU fusion. See Section~\ref{subsec:realworld-eval}.
\end{itemize}

Unlike vision-only or end-to-end policies, GRIP is designed for scenarios with occluded goals, multi-room layouts, cluttered spaces, and incomplete sensory data.

\subsection{Core Modules}

All GRIP variants share a unified planning backbone composed of four key modules:

\begin{enumerate}
    \item \textit{Dynamic Scene Representation:} An open-vocabulary symbolic graph (DovSG) maintains detected anchors and inferred relations, incrementally updated from YOLOv8 detections and geometric estimates (depth or LiDAR) to support reasoning beyond the agent’s current view.

    \item \textit{Symbolic Relay Planning:} A co-occurrence knowledge graph enables anchor chaining (e.g., ``microwave $\rightarrow$ counter $\rightarrow$ fridge'') when goals are occluded. GRIP-F and GRIP-R allow real-time substitutions based on feedback or failures.

    \item \textit{Spatial Path Planning:} GRIP constructs a 2D occupancy grid using RGB-D (simulation) or LiDAR+IMU (real-world). A*/D* planners generate adaptive, obstacle-aware paths to symbolic anchors.

    \item \textit{LLM-Based Introspection:} In GRIP-F and GRIP-R, GPT-4o triggers upon failure (e.g., unreachable anchors), analyzing symbolic traces and scene history to revise anchor chains.
\end{enumerate}

\begin{figure*}[ht]
    \centering
    \fbox{\includegraphics[width=\textwidth]{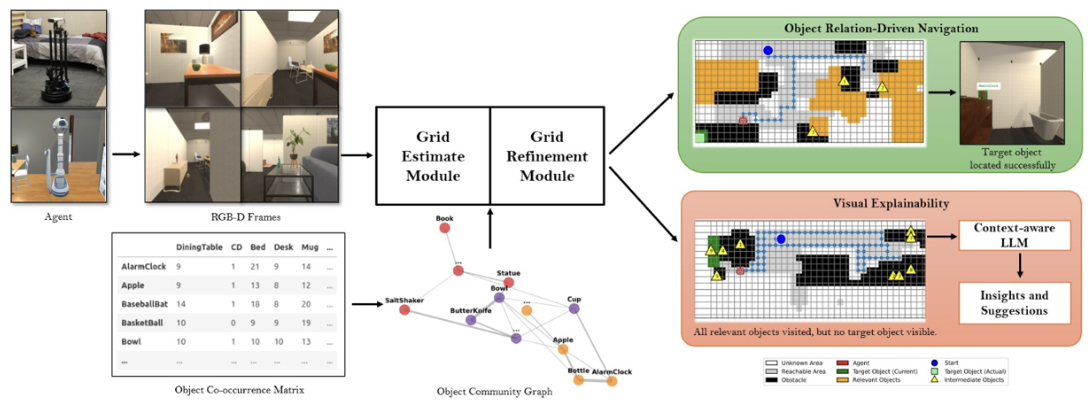}}
    \caption{
    \textit{GRIP-F Architecture.}
    The full symbolic configuration integrates RGB(-D) input, dynamic memory, grid-based planning, symbolic chaining, and LLM recovery, supporting robust navigation under occlusion and ambiguity.
    }
    \label{fig:grip_framework}
\end{figure*}

Figure~\ref{fig:grip_framework} shows GRIP-F—the full symbolic variant—integrating perception, symbolic memory, spatial planning, and LLM-based introspection into a unified loop. GRIP-F supports complex tasks in RoboTHOR, while GRIP-L and GRIP-R simplify modules for simulation efficiency or hardware constraints.

\subsection{Dynamic Scene Representation with DovSG}

The \textit{Dynamic Open-Vocabulary Scene Graph} (DovSG) acts as GRIP's evolving symbolic memory, capturing semantic and spatial aspects of the environment. By maintaining an interpretable graph of entities and their relations, DovSG enables long-horizon reasoning, open-vocabulary grounding, and symbolic planning under partial observability.

\paragraph{Graph Definition}
At each timestep $t$, the environment is encoded as a directed attributed graph:
\[
\mathcal{G}_t = (\mathcal{V}_t, \mathcal{E}_t, \mathcal{A}_t)
\]
where:
\begin{itemize}
    \item $\mathcal{V}_t = \{v_i^t\}_{i=1}^{N_t}$ is the set of nodes (e.g., objects, regions, surfaces),
    \item $\mathcal{E}_t = \{e_{ij}^t\}$ denotes edges representing semantic or spatial relationships,
    \item $\mathcal{A}_t$ includes node and edge attributes (labels, positions, embeddings).
\end{itemize}

\paragraph{Node Representation}
Each node $v_i^t$ is described as:
\[
v_i^t = (l_i, \mathit{f}_i, \mathit{p}_i, \mathit{b}_i)
\]
where $l_i$ is an open-vocabulary label, $\mathit{f}_i \in \mathbb{R}^d$ is a semantic embedding, $\mathit{p}_i \in \mathbb{R}^3$ is the estimated 3D position, and $\mathit{b}_i$ is the bounding box.

\paragraph{Edge Representation}
Each edge $e_{ij}^t$ captures symbolic and spatial relationships:
\[
e_{ij}^t = (\text{rel}_{ij}, \rho_{ij})
\]
with $\text{rel}_{ij}$ as a symbolic relation (e.g., \texttt{on-top-of}, \texttt{next-to}) and $\rho_{ij}$ as a spatial affinity score:
\[
\rho_{ij} = \exp\left(-\frac{\|\mathit{p}_i - \mathit{p}_j\|^2}{\sigma^2}\right) \cdot \cos(\theta_{ij})
\]
where $\theta_{ij}$ is the angular offset between orientations.

\paragraph{Incremental Graph Update}
At each frame, detected nodes are matched to existing ones via:
\[
\text{sim}(v_i^{t-1}, v_j^t) = \lambda \cdot \langle \mathit{f}_i^{t-1}, \mathit{f}_j^t \rangle + (1 - \lambda) \cdot \mathbb{I}[\|\mathit{p}_i^{t-1} - \mathit{p}_j^t\| < \epsilon]
\]
Matched entities are updated, unmatched ones added, and stale nodes gradually pruned.

\paragraph{Language Grounding}
To interpret a language query $q$, GRIP selects the most relevant node $v^*$ using:
\[
v^* = \arg\max_{v_j \in \mathcal{V}_t} \, \text{Score}(v_j \mid q, \mathcal{G}_t)
\]
\[
\text{Score}(v_j) = \alpha \cdot \text{Sim}(q, l_j) + (1 - \alpha) \cdot \max_{e_{kj} \in \mathcal{E}_t} \rho_{kj} \cdot \mathbb{I}[\text{rel}_{kj} \sim q]
\]
This combines label similarity with contextual graph structure.

\paragraph{Functional Role in GRIP}
DovSG enables:
\begin{itemize}
    \item Subgoal decomposition via semantic anchor inference,
    \item Relay chaining from symbolic co-occurrence,
    \item Failure tracing and reasoning via graph traversal,
    \item Language-grounded spatial planning and decision-making.
\end{itemize}

Overall, DovSG bridges perception, language, and geometry, forming the symbolic core of GRIP for persistent and interpretable scene understanding.

\subsection{Symbolic Anchoring and Relay Discovery}
\label{subsec:symbolic-relay}

To navigate in partially observable environments where target objects may be occluded or unreachable, GRIP leverages a \textit{symbolic co-occurrence knowledge graph} to identify intermediate \textit{relay objects}. These anchors enable the agent to plan interpretable, goal-directed trajectories using semantic associations rather than direct visual input alone.

\paragraph{Co-occurrence Knowledge Graph}
Let $G = (V, E, W)$ be an undirected, weighted graph where:
\begin{itemize}
    \item $V = \{o_1, o_2, \dots, o_n\}$ represents object categories,
    \item $E \subseteq V \times V$ denotes observed co-occurrences,
    \item $W: E \rightarrow \mathbb{R}_{\geq 0}$ assigns normalized edge weights.
\end{itemize}

Weights are computed by:
\[
w(o_i, o_j) = \frac{\text{co\_freq}(o_i, o_j)}{\max_{k, \ell} \text{co\_freq}(o_k, o_\ell)}
\]
where $\text{co\_freq}(o_i, o_j)$ counts the number of scenes in which $o_i$ and $o_j$ co-occur.

\paragraph{Semantic Clustering}
To reveal latent semantic communities, we apply modularity-based clustering, partitioning $G$ into subgraphs $\{G_1, G_2, \dots, G_K\}$ with dense intra-cluster connections. Modularity is defined as:
\[
Q = \frac{1}{2m} \sum_{i,j} \left(w(o_i,o_j) - \frac{d_i d_j}{2m}\right) \delta(c_i, c_j)
\]
where $d_i$ is the degree of $o_i$, $m$ is the total weight of all edges, and $\delta(c_i, c_j) = 1$ if $o_i$ and $o_j$ belong to the same cluster.

\paragraph{Relay Trajectory Optimization}
Given a start anchor $o_s$ and a goal object $o_g$, GRIP selects an optimal sequence of relay objects $\mathcal{T} = \{o_{r_1}, \dots, o_{r_m}\}$ by maximizing semantic coherence:
\[
\mathcal{T}^* = \arg\max_{\mathcal{T}} \left[ w(o_s, o_{r_1}) + \sum_{i=1}^{m-1} w(o_{r_i}, o_{r_{i+1}}) + w(o_{r_m}, o_g) \right]
\]
This relay chain serves as a symbolic path when the goal is not directly observable.

\paragraph{LLM-Assisted Subgoal Reasoning}
In cases of sparse or ambiguous co-occurrence priors, we query a Large Language Model (LLM) for plausible subgoal suggestions. Given the current scene graph $\mathcal{G}_t$ and goal $o_g$, the LLM outputs:
\[
\text{RelayCandidates}(o_g) = \{o_1, o_2, \dots, o_k\}
\]
ranked by contextual relevance $P(o_i \mid o_g, \text{scene}, \text{layout})$. Final relay scores are computed by:
\[
\text{FinalScore}(o_i) = \beta \cdot \text{LLMScore}(o_i) + (1 - \beta) \cdot w(o_i, o_g)
\]
where $\beta \in [0,1]$ balances language priors with empirical graph structure.

\paragraph{Integration with Navigation}
The resulting anchor sequence $\mathcal{T}^*$ is fed to the spatial planner. Each anchor is detected via vision-language grounding and reached using grid-based planning. If an anchor is occluded or unreachable, GRIP triggers replanning, selecting the next-best candidate from \texttt{RelayCandidates} or invoking LLM-based hallucination.

This symbolic-augmented strategy enables GRIP to generalize beyond visual input, supporting robust planning under uncertainty while maintaining interpretability.


\subsection{Semantic-Grid Spatial Planning}
\label{subsec:semantic_grid_planning}

To support spatially grounded navigation, GRIP maintains a dynamic \textit{semantic occupancy grid} $\mathcal{M}_t \in \mathbb{R}^{H \times W \times C}$ at each timestep $t$, where:
\begin{itemize}
    \item $H \times W$ defines the 2D spatial resolution of the environment,
    \item $C$ denotes semantic layers (e.g., free space, obstacles, object categories, anchors).
\end{itemize}

Each grid cell $\mathcal{M}_t(x, y)$ encodes a multi-hot vector.
\[
\mathcal{M}_t(x, y) = [\mathbb{1}_{\text{free}}, \mathbb{1}_{\text{obstacle}}, \mathbb{1}_{o_1}, \dots, \mathbb{1}_{o_K}]
\]
where $\mathbb{1}_{\cdot}$ indicates the presence of a specific semantic feature at location $(x, y)$.

\paragraph{Grid Construction and Update}
The occupancy grid is constructed from:
\begin{enumerate}[label=(\alph*)]
    \item RGB-D or RGB observations,
    \item Object detections from DovSG,
    \item Depth-aligned projections registered to a global frame.
\end{enumerate}

Grid updates follow:
\[
\mathcal{M}_t = \text{Update}(\mathcal{M}_{t-1}, \text{DovSG}_t, \text{Pose}_t)
\]
This function integrates current perception, decays outdated entries, and fuses semantic and geometric information to reflect environmental changes over time.

\paragraph{Subgoal Reachability Verification}
For each symbolic relay anchor $o_r \in \mathcal{T}$, GRIP checks whether a reachable free-space cell lies within radius $\epsilon$ of the anchor’s projected location $\mathit{p}_{o_r}$:
\[
\text{Reachable}(o_r) =
\begin{cases}
1, & \exists (x, y) \in \mathcal{M}_t^{\text{free}} : \| (x, y) - \mathit{p}_{o_r} \| \leq \epsilon \\
0, & \text{otherwise}
\end{cases}
\]
Unreachable anchors are deprioritized or removed from the planning sequence.

\paragraph{Path Planning with A*/D*}
If $o_r$ is reachable, GRIP generates a path $\pi_t = \{s_1, \dots, s_T\}$ using A* or D* search over $\mathcal{M}_t^{\text{free}}$:
\[
\pi_t = \text{Planner}(\text{AgentPos}_t, \mathit{p}_{o_r}, \mathcal{M}_t^{\text{free}})
\]
The planned path minimizes navigation cost while maximizing semantic utility, accounting for dynamic obstacles and anchor proximity. The agent executes movement toward $s_{t+1}$ and updates the plan as necessary.

\paragraph{Semantic Replanning and Adaptation}
GRIP continuously adapts the grid and re-evaluates subgoals based on:
\begin{itemize}
    \item Newly detected obstacles or deviation from path $\pi_t$,
    \item Updated symbolic anchors from DovSG,
    \item Environmental changes such as revealed paths or cleared occlusions.
\end{itemize}

This adaptive mechanism ensures robust navigation in dynamic, cluttered, and partially observable environments by incrementally refining $\mathcal{M}_t$ and adjusting plans in real time.


\subsection{Subgoal Decomposition and Behavioral Cloning}
\label{subsec:subgoal_cloning}

To support interpretable long-horizon navigation, GRIP incorporates a transformer-based \textit{behavioral cloning} model that performs symbolic subgoal decomposition. The model learns to map natural language instructions and scene understanding into a sequence of high-level semantic actions by imitating expert demonstrations.

\paragraph{Problem Setup}
Given:
\begin{itemize}
    \item a natural language instruction $\mathcal{I}$,
    \item a dynamic scene graph $\mathcal{G}_t$ generated by DovSG,
    \item and a co-occurrence prior matrix $\mathcal{R} \in \mathbb{R}^{|O| \times |O|}$ encoding statistical object affinities,
\end{itemize}
the model predicts a sequence of symbolic subgoals:
\[
\hat{\mathcal{S}} = \left\{(a_1, o_1), (a_2, o_2), \dots, (a_T, o_T)\right\},
\]
where $a_t$ denotes a symbolic action (e.g., \texttt{goto}, \texttt{inspect}, \texttt{interact}) and $o_t$ an object anchor.

\paragraph{Behavioral Cloning Objective}
The model is trained through supervised imitation learning to maximize the likelihood of expert-annotated subgoal sequences $\mathcal{S}^*$:
\[
\mathcal{L}_{\text{BC}} = -\sum_{t=1}^{T} \log P_\theta\left((a_t^*, o_t^*) \mid \mathcal{I}, \mathcal{G}_t, \mathcal{R}\right),
\]
where $P_\theta$ is implemented as a transformer encoder-decoder with cross-attention over instruction tokens and graph embeddings.

\paragraph{Incorporating Co-occurrence Priors}
To reinforce semantic structure, GRIP fuses the learned distribution with prior knowledge:
\[
P_{\text{fused}}((a, o)) = \lambda \cdot P_\theta((a, o)) + (1 - \lambda) \cdot \mathcal{R}(o, o_g),
\]
where $o_g$ is the target object class and $\lambda \in [0, 1]$ controls the trade-off between model confidence and statistical regularity. This encourages the model to prioritize anchors with high contextual affinity (e.g., selecting \textit{Desk} before \textit{Laptop}).

\paragraph{Subgoal Validation Pipeline}
Each predicted subgoal $(a_t, o_t)$ is passed through a three-stage filter:
\begin{itemize}
    \item \textit{Scene Grounding:} Verify that $o_t$ exists in $\mathcal{G}_t$ with sufficient detection confidence;
    \item \textit{Spatial Feasibility:} Confirm $\text{Reachable}(o_t) = 1$ in the semantic occupancy grid $\mathcal{M}_t$;
    \item \textit{Redundancy Filtering:} Remove duplicate or semantically equivalent anchors (e.g., \textit{Chair} vs. \textit{Stool}).
\end{itemize}
Validated subgoals are forwarded to the spatial planning module for trajectory generation and execution.

\paragraph{Advantages}
This hybrid symbolic-learning framework offers several key benefits. First, it enables dynamic subgoal chaining in cluttered or partially observable scenes, allowing agents to adapt their plans based on environmental complexity. Second, it supports zero-shot generalization to novel objects through open-vocabulary grounding, improving flexibility in unfamiliar settings. Finally, it produces interpretable execution traces that align with human-understandable object semantics, enhancing transparency and explainability in decision-making.


\subsection{Goal-Conditioned Reinforcement Learning (GCRL)}
\label{subsec:gcrl}

To convert symbolic subgoals into continuous low-level control, GRIP incorporates a \textit{Goal-Conditioned Reinforcement Learning} (GCRL) module. This module learns a policy that adapts to semantic targets while optimizing safe and efficient movement through complex 3D environments.

\paragraph{Policy Formulation}
Each symbolic subgoal $g_i = (a_i, o_i)$ is embedded into a latent goal representation $z_{g_i} = \phi(g_i)$, encoding its semantic and functional context. The goal-conditioned policy is expressed as:
\[
\pi_\theta(a_t \mid s_t, z_{g_i}),
\]
where:
\begin{itemize}
    \item $s_t$ denotes the agent’s state at time $t$, composed of:
    \begin{itemize}
        \item RGB-D input $\mathcal{I}_t$,
        \item Egocentric pose $\mathit{p}_t$,
        \item Symbolic object features $\mathcal{F}_t$ from DovSG,
        \item Local semantic occupancy map $\mathcal{M}_t$.
    \end{itemize}
    \item $a_t$ is a discrete low-level action (e.g., \texttt{forward}, \texttt{turn}, \texttt{interact}).
\end{itemize}

\paragraph{Reward Design}
The reward function $R_t$ balances semantic success, spatial efficiency, and safety:
\[
R_t = \lambda_1 R_{\text{goal}} + \lambda_2 R_{\text{progress}} - \lambda_3 R_{\text{collision}} - \lambda_4 R_{\text{drift}},
\]
where:
\begin{itemize}
    \item $R_{\text{goal}}$: Binary reward for successful subgoal completion.
    \item $R_{\text{progress}}$: Dense reward based on reduction in distance to target $o_i$.
    \item $R_{\text{collision}}$: Penalty for collisions or invalid actions.
    \item $R_{\text{drift}}$: Penalty for deviating from the intended semantic goal (e.g., heading toward unrelated objects).
\end{itemize}
The coefficients $\lambda_1, \dots, \lambda_4$ control the trade-off between goal-directedness, safety, and semantic coherence.

\paragraph{Optimization and Curriculum Strategy}
The policy $\pi_\theta$ is optimized using Proximal Policy Optimization (PPO), an on-policy reinforcement learning algorithm known for sample efficiency and training stability. To promote generalization and robustness, we employ curriculum learning:
\begin{enumerate}[label=(\alph*)]
    \item Start with simple layouts and short-horizon subgoals.
    \item Gradually introduce occlusions, distractors, and longer semantic chains.
    \item Resample difficult subgoals based on failure statistics to focus training on weak points.
\end{enumerate}

\paragraph{Benefits}
The GCRL module enables:
\begin{itemize}
    \item \textit{Subgoal-adaptive behavior:} Action policies are tailored to specific symbolic intents.
    \item \textit{Scene-aware control:} Combines semantic and spatial observations for informed decision-making.
    \item \textit{Robust execution:} Maintains performance under occlusion, dynamic layouts, and sensory noise.
\end{itemize}

Overall, GCRL bridges symbolic abstraction and continuous control, completing GRIP’s end-to-end pipeline from semantic reasoning to real-world execution.

\subsection{LLM-Based Introspection and Recovery}
\label{subsec:llm-recovery}

To improve resilience in failure scenarios—such as occlusions, unreachable anchors, or semantic misalignment—GRIP integrates a \textit{Large Language Model (LLM)-driven introspection and recovery module}. This component reanalyzes symbolic navigation failures and proposes revised plans through contextual and commonsense reasoning.

\paragraph{Failure Diagnosis}
Upon detecting a failure (e.g., prolonged search, anchor loop, or unreachable target), GRIP generates a structured prompt $\mathcal{P}_t$ encoding:
\begin{itemize}
    \item The sequence of visited anchors $\mathcal{T}_{\text{visited}} = \{o_1, o_2, \dots, o_k\}$,
    \item The current symbolic scene graph $\mathcal{G}_t$, including object labels, relations, and spatial states,
    \item Diagnostic cues such as repetitive trajectories, unresolved goals, or mismatches between expected and observed objects.
\end{itemize}

This prompt is submitted to an LLM (e.g., GPT-4, Gemini) via a task-specific query:
\[
\hat{o}_r, \hat{\mathcal{T}} \leftarrow \text{LLM}(\mathcal{P}_t),
\]
where $\hat{o}_r$ is a proposed alternative relay anchor, and $\hat{\mathcal{T}}$ is a revised anchor sequence. These suggestions are filtered for spatial feasibility using the current semantic grid $\mathcal{M}_t$ before adoption.

\paragraph{Recovery Execution}
Once the revised subgoals are validated, they are seamlessly integrated into GRIP’s symbolic planner. Rather than resetting the entire episode, the introspection module performs a soft reset—preserving the agent’s current navigation context while refreshing the symbolic anchor set. The goal-conditioned policy is then reinitialized with the updated trajectory, allowing the agent to resume navigation without restarting or re-executing prior steps. This design ensures continuity and minimizes disruptions during online replanning.

\paragraph{Advantages}
The LLM-based recovery mechanism introduces several significant benefits. First, it supports robust replanning in scenarios where co-occurrence priors or symbolic chains fail, allowing the system to adapt on-the-fly. Second, it incorporates commonsense reasoning by leveraging linguistic priors and external world knowledge that are not explicitly encoded in the symbolic graph. Third, it enhances failure tolerance by enabling dynamic revision of symbolic intents and action sequences, thereby avoiding hard resets.

Together, this introspective module effectively bridges the gap between perception, planning, and reasoning. By doing so, it enables GRIP to operate reliably even in ambiguous, cluttered, or partially observable environments.

\subsection{Summary}
\label{subsec:grip-summary}

GRIP is a modular, interpretable navigation framework with three variants: \textit{GRIP-F} (Full), \textit{GRIP-L} (Lightweight), and \textit{GRIP-R} (Real-World). These models balance symbolic reasoning, spatial planning, and language-guided introspection to support robust, explainable navigation across varying environments.

\begin{itemize}
    \item \textit{Dynamic Open-Vocabulary Scene Graph (DovSG)}—core to all variants—converts RGB-D input into symbolic scene graphs for grounded perception and subgoal inference.

    \item \textit{Co-occurrence Knowledge Graph}—in GRIP-F and GRIP-R—infers intermediate anchors when targets are occluded or absent.

    \item \textit{Semantic Occupancy Grid}—shared by GRIP-L and GRIP-R—merges geometry and semantics to guide goal-conditioned planning.

    \item \textit{Goal-Conditioned Reinforcement Learning (GCRL)}—exclusive to GRIP-F—maps symbolic goals to continuous actions in complex simulations.

    \item \textit{LLM-Based Introspection and Recovery}—used in GRIP-F and optionally GRIP-R—supports replanning through language-guided symbolic reasoning.
\end{itemize}

\textit{GRIP-F} is tailored for complex settings like RoboTHOR;  
\textit{GRIP-L} serves as a lightweight variant for simpler AI2-THOR tasks;  
\textit{GRIP-R} deploys these capabilities on physical robots with real-time sensing and control.  
Together, these variants enable resilient and interpretable navigation in both simulation and real-world domains.

\section{Real-World Robot Implementation: GRIP-R}
\label{sec:realworld}

We deploy \textit{GRIP-R}, the real-world extension of GRIP, on a mobile robot navigating indoor environments with occlusions, layout changes, and partial observability. GRIP-R integrates symbolic reasoning, spatial planning, and vision-language grounding for interpretable, goal-driven navigation.

\subsection{GRIP-R Framework Overview}
\label{subsec:grip-overview}

GRIP-R incorporates four core modules:

\begin{enumerate}
    \item \textit{Dynamic Scene Representation:} Builds an open-vocabulary scene graph (DovSG) from RGB-D input for symbolic memory and semantic context.
    \item \textit{Symbolic Relay Planning:} Uses a co-occurrence knowledge graph to infer subgoals when targets are unobserved.
    \item \textit{Spatial Path Planning:} Maintains a real-time 2D occupancy grid for A*/D* trajectory planning.
    \item \textit{LLM-Based Introspection:} Employs a language model for replanning based on symbolic context.
\end{enumerate}

\subsection{Physical Robot Platform}
\label{subsec:hardware}

GRIP-R is built on the Jetbot ROS platform and equipped with:

\begin{itemize}
    \item \textit{Jetson Nano:} Runs YOLO and GRIP-R modules in real time.
    \item \textit{MPU-9250 IMU:} Tracks heading and validates motion.
    \item \textit{RPLIDAR A1:} Builds 2D occupancy grids.
    \item \textit{Sony IMX219 RGB Camera:} Captures visual input for object detection and scene graphs.
    \item \textit{RP2040 Microcontroller:} Manages low-level control and sensor I/O.
\end{itemize}

\subsection{Core Software and Algorithms}
\label{subsec:algorithms}

GRIP-R fuses real-time perception, symbolic abstraction, and motion planning via the following components:

\begin{itemize}
    \item \textit{Object Detection and Symbolic Grounding:} 
    YOLO detects objects from RGB frames. Detected bounding boxes are converted into symbolic triples and inserted into the Dynamic Open-Vocabulary Scene Graph (DovSG) for high-level planning.

    \item \textit{Motion Primitives and IMU Validation:} 
    The robot executes discrete actions: \textit{Move Forward (10 cm)}, \textit{Rotate Left (90°)}, and \textit{Rotate Right (90°)}. Each action is confirmed using MPU-9250 IMU feedback to validate displacement and heading.

    \item \textit{Heading and State Estimation:} 
    IMU-derived heading and displacement data track robot pose within a global grid, maintaining alignment between symbolic targets and physical state.

    \item \textit{Occupancy Grid Mapping:} 
    RPLIDAR constructs a 2D occupancy map in real time, marking static objects, dynamic obstacles, and unexplored space to support safe path planning.

    \item \textit{Path Planning and Replanning:} 
    A* or D* algorithms are used to compute collision-free paths from the current location to symbolic goals. Replanning is triggered by unexpected obstacles, occlusions, or updated subgoals.

    \item \textit{Target Localization via RGB–LIDAR Fusion:} 
    Combines RGB-based detection, IMU heading, and LIDAR depth to estimate object coordinates on the grid. Accuracy improves with centered and proximate targets.
\end{itemize}

\vspace{0.5em}
\noindent
The following pseudocode outlines the symbolic-grounded navigation process in GRIP-R:

\begin{algorithm}[H]
\caption{GRIP-R Symbolic Navigation Loop}
\label{alg:grip-r}
\begin{algorithmic}[1]
\State \textbf{Input:} RGB-D stream, LIDAR scans, IMU readings, target instruction
\State Initialize scene graph $\mathcal{G} \leftarrow \emptyset$, occupancy grid $\mathcal{M}$
\While{task not complete}
    \State Detect objects $\mathcal{O} \leftarrow$ YOLO$(\text{RGB})$
    \State Update DovSG $\mathcal{G} \leftarrow \mathcal{G} \cup \text{Ground}(\mathcal{O})$
    \State Estimate pose $p_t \leftarrow$ IMU$(\cdot)$
    \State Update grid $\mathcal{M} \leftarrow$ LIDAR$(\cdot)$
    \State Plan path $\pi \leftarrow$ A*/D*$(p_t, \text{Goal}(\mathcal{G}))$
    \For{each action $a \in \pi$}
        \State Execute $a$, validate with IMU
        \If{obstacle $\vee$ failed action}
            \State Replan $\pi \leftarrow$ D*$\!$(updated $\mathcal{M}$, $\mathcal{G}$)
        \EndIf
    \EndFor
\EndWhile
\end{algorithmic}
\end{algorithm}

\section{Results and Evaluation}
\subsection{Datasets and Environments}
\label{subsec:datasets}

We evaluate GRIP across three setups:
(1) \textit{GRIP-L} in AI2-THOR for lightweight symbolic reasoning,
(2) \textit{GRIP-F} in RoboTHOR for full-stack semantic policy learning,
(3) \textit{GRIP-R} for real-world deployment on a Jetbot Pro robot.
This setup assesses GRIP’s generalizability, robustness, and sim-to-real transfer.

\subsubsection*{AI2-THOR: GRIP-L Evaluation}
AI2-THOR~\cite{kolve2017ai2} provides \textit{120 photorealistic scenes} across four room types and 100+ objects. GRIP-L uses the \textit{Dynamic Open-Vocabulary Scene Graph} and \textit{Semantic Occupancy Grid} for symbolic planning.
We follow the official split (75 train / 15 val / 30 test). AI2-THOR’s layout and clear object visibility enable efficient symbolic chaining with minimal perceptual noise.

\subsubsection*{RoboTHOR: GRIP-F Evaluation}
RoboTHOR~\cite{deitke2020robothor} features \textit{89 cluttered apartment-style scenes} with occlusion and nested objects. Using the AllenAct ObjectNav split (60 / 16 / 13), GRIP-F applies the Co-occurrence Knowledge Graph, Goal-Conditioned RL, and LLM-based introspection to handle ambiguity and long-horizon reasoning.

\subsubsection*{Real-World Apartment: GRIP-R Evaluation}
GRIP-R is deployed on a Jetbot Pro in a real apartment with three rooms and five diverse object targets (e.g., TV, Backpack). Equipped with YOLOv8 (RGB), 2D LiDAR, and an IMU, it performs symbolic chaining, localization, and introspection under occlusion, lighting changes, and layout shifts—demonstrating strong sim-to-real transfer.

\subsubsection*{Task Definition: ObjectNav}
All variants are evaluated on Object-Goal Navigation, where the agent receives a textual goal (e.g., “find the backpack”) and must locate the object from a random start point, handling both visible and occluded scenarios.

\subsubsection*{Environment Comparison}
Table~\ref{tab:dataset_comparison} compares GRIP-L, GRIP-F, and GRIP-R settings in terms of environment features, object characteristics, and model components.

\begin{table*}[ht!]
\centering
\caption{Overview of GRIP Deployment in AI2-THOR, RoboTHOR, and Real-World Environments}
\label{tab:dataset_comparison}
\footnotesize
\begin{tabular}{l|c|c|c}
\hline
\textbf{Feature} & \textbf{AI2-THOR (GRIP-L)} & \textbf{RoboTHOR (GRIP-F)} & \textbf{Real-World (GRIP-R)} \\
\hline
\textit{Scenes / Room Types} & 120 / Kitchen, Bedroom, etc. & 89 / Multi-room Apts. & 1 / Kitchen, Bedroom, Living \\
\textit{Object Categories} & 100+ (e.g., Mug, Laptop) & 100+ (e.g., TV, Microwave) & 5 real (e.g., TV, Fridge) \\
\textit{Object Placement} & Structured / Visible & Nested / Occluded & Natural / Passive \\
\textit{Observability} & RGB-D (Synthetic) & RGB-D (Simulated) & RGB + LiDAR + IMU \\
\textit{Depth Estimation} & Ground-truth & Simulated & LiDAR + Angle Inference \\
\textit{Data Split} & 75 / 15 / 30 & 60 / 16 / 13 & N/A (Single Run) \\
\textit{Primary Role} & Symbolic Planning & Semantic + Policy Learning & Real Deployment \\
\textit{GRIP Modules Used} & DovSG, Grid & DovSG, KG, GCRL, LLM & DovSG, Grid, KG, LLM \\
\hline
\end{tabular}
\end{table*}

\subsection{Evaluation Metrics}
\label{subsec:measures}

We evaluate GRIP using standard embodied AI metrics that assess task completion, spatial efficiency, and semantic action optimization:

\begin{itemize}
    \item \textit{Success Rate (SR)}~($\uparrow$): Percentage of episodes where the agent stops within 1 meter of the target:
    \[
    \text{SR} = \frac{1}{N} \sum_{i=1}^{N} \mathbb{I}\left[d(g_i, p_i^{\text{stop}}) \leq 1.0 \right]
    \]
    where $N$ is the number of episodes, $g_i$ is the goal location, and $p_i^{\text{stop}}$ is the agent’s final position.

   \item \textit{Success-weighted Path Length (SPL)}~($\uparrow$): Efficiency ratio between shortest and actual path, conditional on success:
    \[
    \text{SPL} = \frac{1}{N} \sum_{i=1}^{N} \frac{\mathbb{I}_i \cdot \ell_i}{\max(p_i, \ell_i)}
    \]
   where $\ell_i$ is the shortest path, $p_i$ is the path taken, and $\mathbb{I}_i$ indicates success.

    \item \textit{Success-weighted Action Efficiency (SAE)}~($\uparrow$): Like SPL but factors in semantic action cost (e.g., \texttt{Pickup}, \texttt{Open}):
    \[
    \text{SAE} = \frac{1}{N} \sum_{i=1}^{N} \frac{\mathbb{I}_i \cdot \ell_i}{\max(a_i, \ell_i)}
    \]
    where $a_i$ is the count of high-level actions in episode $i$.
\end{itemize}

Metrics are computed under two evaluation regimes:

\begin{itemize}
    \item \textit{ALL:} Includes all episodes across short and long trajectories.
    \item \textit{$L \geq 5$:} Focuses on episodes where the start-to-goal geodesic distance exceeds 5 steps, emphasizing long-horizon semantic reasoning.
\end{itemize}

These metrics jointly measure GRIP’s ability to navigate accurately, efficiently, and semantically—critical for generalizable, interpretable agents.

\subsection{Baseline Methods}
\label{subsec:baselines}

We evaluate GRIP against a comprehensive suite of baseline methods organized into four main categories that correspond directly with the experimental tables: Transformer Navigation, Object-Goal Navigation, Semantic Dynamics, and Hierarchical RL and Planning.

\paragraph*{(1) Transformer Navigation}
Transformer-based methods integrate multi-modal vision-language models and semantic scene priors with transformer architectures to enhance semantic reasoning, long-horizon planning, and generalization. These methods achieve state-of-the-art performance, especially in large-scale benchmarks, but can be computationally expensive or brittle in novel configurations.

\begin{itemize}
    \item \textit{HOZ$_{i}$, HOZ$_{i++}$, HOZ$_{e++}$}~\cite{zhang2025hoz++}: Object–zone transformers with hierarchical scene reasoning.
    \item \textit{AKGVP, AKGVP-CI}~\cite{xu2024aligning}: Vision–language alignment with symbolic goal supervision.
    \item \textit{CGI-GAIL}~\cite{meng2025context}: TransH-based symbolic scene graphs with generative imitation learning.
    \item \textit{MT}~\cite{dang2023multiple}: Meta-learning and dense attention mechanisms over trajectory embeddings.
    \item \textit{Zheng et al.}~\cite{zheng2024two}: Object-view tokenization with dual-stage reasoning over transformer backbones.
\end{itemize}

\paragraph*{(2) Object-Goal Navigation}
These RL-based models operate in an end-to-end fashion from egocentric observations to actions, often without explicit memory or semantics. While strong in short-horizon seen tasks, they struggle in occluded or long-horizon cases.

\begin{itemize}
    \item \textit{SP}, \textit{SA}~\cite{chaplot2020object, mayo2021visual}: Use end-to-end or meta-RL for visual goal completion.
    \item \textit{EOTP}, \textit{SpAtt}~\cite{mayo2021visual}: Introduce attention priors for goal localization.
    \item \textit{ORG}, \textit{ORG+TPN}~\cite{chaplot2020object, du2020learning}: Add topological recovery and object-relational memory.
\end{itemize}

\paragraph*{(3) Semantic Dynamics}
These models incorporate semantic priors or spatial reasoning to improve long-horizon object search. They perform well in static scenes but are less effective in dynamically changing or highly cluttered layouts.

\begin{itemize}
    \item \textit{SSCNav}~\cite{liang2021sscnav}: Combines semantic scene completion with object-goal control.
    \item \textit{IOM}~\cite{xie2023implicit}: Constructs implicit object maps through visual memory.
    \item \textit{L-sTDE}~\cite{zhang2023layout}: Adds layout estimation to assist in long-horizon planning.
    \item \textit{VTNet}~\cite{du2021vtnet}: Uses transformer-based spatial attention for goal prediction.
    \item \textit{CRG}~\cite{hu2024building}: Learns compositional relational graphs for symbolic reasoning over affordances.
\end{itemize}

\paragraph*{(4) HRL and Planning}
Hierarchical methods use modular planners or symbolic policies to compose subgoals, often relying on structural decomposition. They offer transparency and robustness but are limited in generalization without high-level semantic guidance.

\begin{itemize}
    \item \textit{ORG+TPN}~\cite{du2020learning}: Employs topological planning networks for recovery and high-level control.
    \item \textit{EOTP}~\cite{mayo2021visual}: Combines episodic task priors with subgoal learning.
\end{itemize}

\paragraph*{(5) GRIP Framework (Ours)}
GRIP generalizes across synthetic and real-world settings by integrating visual semantics, symbolic abstraction, and transformer-guided decision policies.

\begin{itemize}
    \item \textit{GRIP-L} (AI2-THOR): Leverages symbolic goal graphs (DovSG) and semantic relay via transformers.
    \item \textit{GRIP-F} (RoboTHOR): Adds transformer-guided goal disambiguation and planning with symbolic affordance priors.
    \item \textit{GRIP-R} (Real World): Integrates co-occurrence knowledge, language-based recovery, and semantic segmentation for sim-to-real robustness.
\end{itemize}

\vspace{0.5em}
\noindent
\textit{Summary of Strengths and Weaknesses}
\begin{itemize}
    \item \textit{Transformer methods}: High semantic fidelity, strong generalization, but computationally intensive.
    \item \textit{Object-goal RL}: Fast adaptation but poor symbolic grounding.
    \item \textit{Semantic dynamics}: Good spatial modeling, but limited semantic flexibility.
    \item \textit{Hierarchical planning}: Clear interpretability, but limited goal diversity handling.
    \item \textit{GRIP}: Combines semantic reasoning, symbolic planning, and modular execution across all settings.
\end{itemize}

\begin{table*}[ht!]
\footnotesize
\centering
\caption{Performance Comparison on the AI2-THOR Dataset. Evaluation includes Success Rate (SR), Success-weighted Path Length (SPL), and Success-weighted Action Efficiency (SAE), reported for all episodes and long-horizon scenarios ($L \geq 5$). GRIP-L surpasses all baselines in SPL and SAE, particularly excelling in long-horizon planning.}
\label{tab:ai2thor_comparison}
\begin{tabular}{L{3cm}|L{2cm}|C{1.2cm}|C{1.2cm}|C{1.2cm}|C{1.2cm}|C{1.2cm}|C{1.2cm}}
\toprule
\multirow{2}{*}{\bf Category} & \multirow{2}{*}{\bf Method} & \multicolumn{3}{c|}{\textbf{ALL (\%)}} & \multicolumn{3}{c}{\textbf{$L \geq 5$ (\%)}} \\
\cmidrule(lr){3-5} \cmidrule(lr){6-8}
& & SR & SPL & SAE & SR & SPL & SAE \\ 
\midrule

\multirow{9}{*}{\bf Transformer Navigation}
& NavTr~\cite{mao2024navtr} & 72.30 & 19.70 & -- & 59.10 & 23.60 & -- \\
& AKGVP-CI~\cite{xu2024aligning} & 76.78 & 39.63 & -- & 65.45 & 39.01 & -- \\
& CGI-GAIL~\cite{meng2025context} & 77.59 & 46.25 & -- & 69.18 & 46.10 & -- \\
& TDANet~\cite{lian2024tdanet} & 78.20 & 30.60 & -- & 67.00 & 33.40 & -- \\
& CRG-TSR~\cite{hu2024building} & 80.00 & 45.60 & -- & 73.60 & 45.10 & -- \\
& MVVT~\cite{lu2025multi} & 81.37 & 62.28 & 39.45 & 75.78 & 60.02 & 44.59 \\
& HOZ$_{i++}$~\cite{zhang2025hoz++} & 83.14 & 47.72 & 35.62 & -- & -- & -- \\
& MT~\cite{dang2023multiple} & 86.47 & 47.02 & 41.63 & 81.08 & 50.26 & 44.51 \\
& Zheng et al.~\cite{zheng2024two} & \textbf{87.69} & 52.09 & 39.75 & 83.02 & 54.15 & 44.40 \\
\midrule

\multirow{4}{*}{\bf Object-Goal Navigation}
& ORG~\cite{chaplot2020object} & 67.84 & 36.94 & 26.18 & 58.58 & 35.61 & 27.34 \\
& OMT~\cite{fukushima2022object} & 70.54 & 30.43 & 29.16 & 59.78 & 31.63 & 24.37 \\
& SSCNav~\cite{liang2021sscnav} & 76.28 & 24.43 & 26.58 & 68.11 & 29.93 & 24.97 \\
& PONI~\cite{ramakrishnan2022poni} & 79.64 & 35.60 & 28.17 & 73.28 & 36.37 & 29.17 \\
\midrule

\multirow{3}{*}{\bf Semantic Dynamics}
& VTNet~\cite{du2021vtnet} & 72.37 & 45.47 & 30.51 & 64.56 & 40.34 & 28.15 \\
& L-sTDE~\cite{zhang2023layout} & 74.19 & 40.30 & -- & 64.01 & 39.97 & -- \\
& IOM~\cite{xie2023implicit} & 82.99 & 47.40 & 33.04 & 77.95 & 48.78 & 37.69 \\
\midrule

\multirow{2}{*}{\bf HRL and Planning}
& EOTP~\cite{mayo2021visual} & 65.61 & 38.93 & 24.86 & -- & -- & -- \\
& Wang~\cite{wang2025goal} (KK) & 83.73 & 57.03 & -- & -- & -- & -- \\
\midrule

\textbf{GRIP Framework}
& \textbf{GRIP-L (Ours)} 
& \underline{84.10} {\footnotesize (${-}3.59$)} 
& \textbf{63.48} {\footnotesize ($+1.20$)} 
& \textbf{65.13} {\footnotesize ($+23.50$)} 
& \textbf{91.79} {\footnotesize ($+8.77$)} 
& \textbf{93.88} {\footnotesize ($+39.73$)} 
& \textbf{80.30} {\footnotesize ($+35.71$)} \\
\bottomrule
\end{tabular}
\end{table*}

\subsection{Quantitative Results on AI2-THOR (GRIP-L)}
\label{subsec:quant-eval}

Table~\ref{tab:ai2thor_comparison} benchmarks \textit{GRIP-L} against recent ObjectNav models on AI2-THOR, reporting Success Rate (SR), SPL, and SAE for all episodes and for long-horizon tasks ($L \geq 5$).

\vspace{0.5em}
\noindent
\textit{Key Observations:}
\begin{itemize}
    \item \textit{ObjectNav baselines} (e.g., ORG~\cite{chaplot2020object}, OMT~\cite{fukushima2022object}) achieve reasonable SR but lag in efficiency (SPL/SAE) due to poor spatial grounding.
    \item \textit{Semantic dynamics models} like IOM~\cite{xie2023implicit} improve mapping fidelity, but lack compositional goal reasoning.
    \item \textit{Transformer-based models} (e.g., MVVT~\cite{lu2025multi}, MT~\cite{dang2023multiple}) yield strong SR, yet show diminished SPL/SAE on longer tasks.
    \item \textit{Hierarchical methods} (e.g., Wang~\cite{wang2025goal}) exhibit strong SPL but face generalization gaps in unseen scenes.
    \item \textit{GRIP-L (Ours)} achieves top performance across all metrics, especially in long-horizon settings: SR = \textit{91.79\%}, SPL = \textit{93.88\%}, and SAE = \textit{80.30\%}, outperforming the next-best model by +8.8\% SR, +39.7\% SPL, and +35.7\% SAE.
\end{itemize}

\vspace{0.5em}
\noindent
\textit{Why GRIP-L Excels:}
\begin{itemize}
    \item \textit{High-resolution semantic grids} improve precision in path planning.
    \item \textit{Symbolic relay graphs} support interpretable, compositional navigation.
    \item \textit{Lightweight transformers} ensure efficient spatial-symbolic fusion.
\end{itemize}

\noindent
By integrating geometric precision and symbolic abstraction, GRIP-L generalizes well to novel layouts and goals. It sets a new standard for interpretable, scalable ObjectNav, particularly in long-horizon scenarios.

\begin{table*}[ht!]
\footnotesize
\centering
\caption{Performance Comparison on RoboTHOR. Evaluated using Success Rate (SR), Success-weighted Path Length (SPL), and Success-weighted Action Efficiency (SAE) across all episodes and long-horizon cases ($L \geq 5$). GRIP-F consistently outperforms all prior baselines across all metrics.}
\label{tab:robothor_comparison_aligned}
\begin{tabular}{L{3cm}|L{2.2cm}|C{1.2cm}|C{1.2cm}|C{1.2cm}|C{1.2cm}|C{1.2cm}|C{1.2cm}}
\toprule
\multirow{2}{*}{\bf Category} & \multirow{2}{*}{\bf Method} & \multicolumn{3}{c|}{\textbf{ALL (\%)}} & \multicolumn{3}{c}{\textbf{$L \geq 5$ (\%)}} \\
\cmidrule(lr){3-5} \cmidrule(lr){6-8}
& & SR & SPL & SAE & SR & SPL & SAE \\ \hline

\multirow{8}{*}{\bf Transformer Navigation}
& HOZ$_i$~\cite{zhang2025hoz++} & 33.28 & 22.13 & 16.66 & -- & -- & -- \\
& AKGVP~\cite{xu2024aligning} & 39.69 & 25.84 & -- & 28.55 & 18.79 & -- \\
& AKGVP-CI~\cite{xu2024aligning} & 44.53 & 27.61 & -- & 32.68 & 20.55 & -- \\
& CGI-GAIL~\cite{meng2025context} & 48.30 & 28.77 & -- & 36.69 & 21.89 & -- \\
& HOZ$_{i++}$~\cite{zhang2025hoz++} & 52.92 & 29.02 & 30.41 & -- & -- & -- \\
& HOZ$_{e++}$~\cite{zhang2025hoz++} & 57.32 & 31.17 & 29.08 & -- & -- & -- \\
& MT~\cite{dang2023multiple} & 73.86 & 38.46 & 42.44 & 71.85 & 38.74 & 42.52 \\
& Zheng et al.~\cite{zheng2024two} & \underline{76.00} & \underline{40.94} & \underline{42.89} & \underline{75.90} & \underline{40.26} & \underline{43.02} \\
\hline

\multirow{4}{*}{\bf Object-Goal Navigation}
& SP~\cite{chaplot2020object} & 26.22 & 16.90 & 14.22 & -- & -- & -- \\
& SA~\cite{mayo2021visual} & 27.54 & 17.37 & 16.35 & 21.53 & 15.79 & 17.60 \\
& ORG~\cite{chaplot2020object} & 45.16 & 26.47 & 25.45 & -- & -- & -- \\
\hline

\multirow{3}{*}{\bf Semantic Dynamics}
& VTNet~\cite{du2021vtnet} & 47.36 & 28.01 & 26.50 & 30.19 & 17.42 & 18.20 \\
& L-sTDE~\cite{zhang2023layout} & 42.13 & 24.54 & -- & 32.04 & 17.44 & -- \\
& IOM~\cite{xie2023implicit} & 42.07 & 27.47 & 24.42 & 37.26 & 22.23 & 25.04 \\
\hline

\multirow{2}{*}{\bf HRL and Planning}
& EOTP~\cite{mayo2021visual} & 28.84 & 18.82 & 13.88 & -- & -- & -- \\
& ORG+TPN~\cite{du2020learning} & 45.86 & 27.17 & 25.85 & 22.25 & 16.64 & -- \\
\hline

\textbf{GRIP Framework} 
& \textbf{GRIP-F (Ours)} 
& \textbf{85.64} {\footnotesize (+9.64)} 
& \textbf{61.03} {\footnotesize (+20.09)} 
& \textbf{67.06} {\footnotesize (+24.17)} 
& \textbf{82.56} {\footnotesize (+6.66)} 
& \textbf{65.28} {\footnotesize (+25.02)} 
& \textbf{66.93} {\footnotesize (+23.91)} \\
\hline

\end{tabular}
\end{table*}

\subsection{Evaluation on RoboTHOR (GRIP-F)}
\label{subsec:robothor-eval}

Table~\ref{tab:robothor_comparison_aligned} presents the performance of our full model, \textit{GRIP-F}, on the RoboTHOR benchmark—characterized by realistic apartment-style layouts with clutter, occlusions, and semantically complex goal conditions. These environments challenge both generalization and planning under partial observability.

\vspace{0.5em}
\noindent
\textit{Performance Highlights:}
\begin{itemize}
    \item \textit{Object-Goal Navigation} baselines, such as SP, SA, and SAVN, suffer from poor semantic grounding and struggle in long-horizon scenarios.
    \item \textit{Semantic Dynamics} methods (e.g., IOM, VTNet) offer richer environmental priors but degrade under ambiguity or irregular layouts.
    \item \textit{Transformer Navigation} approaches (e.g., CGI-GAIL, HOZ$^\text{e++}$, MT) improve visual-language alignment and success rate, but often yield suboptimal paths.
    \item \textit{GRIP-F (Ours)} achieves the best overall performance: \textit{85.64\% SR}, \textit{61.03\% SPL}, and \textit{67.06\% SAE}, with strong performance in long-horizon tasks: \textit{82.56\% SR}, \textit{65.28\% SPL}, and \textit{66.93\% SAE}.
\end{itemize}

Importantly, these results are obtained \textit{without invoking} GRIP-F’s LLM-based introspection or recovery modules. All tasks are completed via symbolic relay planning and co-occurrence graph inference. Symbolic traces are captured for analysis, but GPT-based feedback is disabled to ensure comparability.

\subsubsection*{Analysis}

RoboTHOR introduces several core challenges:
\begin{itemize}
    \item \textit{Partial Observability:} Hidden or nested goals require multi-hop symbolic inference.
    \item \textit{Cluttered Layouts:} Occlusions and tight spatial configurations stress trajectory planning.
    \item \textit{Semantic Ambiguity:} Visually similar distractors demand context-aware disambiguation.
\end{itemize}

GRIP-F addresses these through:
\begin{itemize}
    \item \textit{Symbolic co-occurrence chaining} across reference objects for robust goal localization.
    \item \textit{Occupancy-aware spatial planning} to ensure compact, semantically valid trajectories.
\end{itemize}

While LLM modules for failure explanation and recovery exist within GRIP-F, they were deliberately excluded in this evaluation to maintain parity with prior work. Future analysis will investigate their contributions to robustness and explainability.

GRIP-F demonstrates state-of-the-art performance on RoboTHOR, validating the strength of symbolic planning and spatial reasoning without relying on neural introspection.

\subsection{Ablation and Object-Level Analysis}
\label{subsec:ablation-object}

\paragraph{Object-Level Robustness (See Appendix~\ref{app:objectwise-eval})}
To assess GRIP’s semantic generalization, we conducted object-wise evaluations across 21 and 13 object categories in AI2-THOR and RoboTHOR, respectively. GRIP-L excelled in visibility-dominant scenarios (e.g., \textit{Sink}, \textit{DeskLamp}), while GRIP-F handled occluded, ambiguous objects (e.g., \textit{Laptop}, \textit{Mug}) via symbolic relay chaining and co-occurrence reasoning. Detailed results are in Appendix~\ref{app:objectwise-eval}.

\paragraph{Component-Level Ablation on RoboTHOR}
We ablated key symbolic modules in GRIP-F:
\begin{itemize}
    \item \textit{Co-occurrence Graph} — anchors goal-related objects for semantic grounding.
    \item \textit{Symbolic Relay Chaining} — enables subgoal decomposition.
    \item \textit{Random Anchors} — tests sensitivity to meaningful anchor selection.
\end{itemize}

\begin{table}[ht!]
\footnotesize
\centering
\caption{Ablation on GRIP-F (RoboTHOR). Each module improves success (SR), efficiency (SPL), and action economy (SAE).}
\label{tab:ablation}
\begin{tabular}{l|c|c|c}
\hline
\textbf{Model Variant} & \textbf{SR (\%)} & \textbf{SPL (\%)} & \textbf{SAE (\%)} \\
\hline
GRIP-F (Full Model) & \textbf{85.64} & \textbf{61.03} & \textbf{67.06} \\
w/o Co-occurrence Graph & 42.56 & 28.93 & 20.41 \\
w/o Symbolic Relay Chaining & 40.00 & 29.08 & 27.70 \\
w/ Random Relay Objects & 34.18 & 23.92 & 18.01 \\
\textit{Static Grid (Oracle)} & 95.38 & 94.69 & 73.01 \\
\hline
\end{tabular}
\end{table}

\paragraph{Findings}
Disabling the co-occurrence graph or relay chaining causes severe performance drops—SR nearly halves—highlighting their necessity for goal inference and trajectory optimization. Random anchor selection further degrades performance, confirming the importance of semantic priors. The oracle static-grid baseline achieves high scores but assumes full map access, unlike GRIP-F’s egocentric mapping.

\paragraph{Failure Recovery and Limitations}
GRIP-F logs symbolic failures for future LLM-based recovery, though introspection is not used in this evaluation. While efficient, GRIP-F’s anchor selection lacks visibility filtering—occluded anchors may be chosen. Depth-based heuristics  may further improve planning feasibility.

\vspace{0.5em}
\noindent
\textit{Takeaway:} GRIP-F’s symbolic stack—co-occurrence priors, relay chaining, and dynamic planning—collectively enables robust and interpretable navigation under uncertainty.

\subsection{From Simulation to Reality: GRIP-R Evaluation}
\label{subsec:realworld-eval}

To validate GRIP’s real-world transferability, we deployed \textit{GRIP-R} on a Jetbot-based mobile robot in a three-room apartment. This environment presents real-world challenges—sensor noise, dynamic lighting, occlusions, and limited actuation—serving as a robust testbed for GRIP’s symbolic planning and relay reasoning.

We evaluated five everyday objects—\textit{Laptop}, \textit{TV}, \textit{Refrigerator}, \textit{Sports Ball}, and \textit{Backpack}—with 15 trials per object. Metrics include Success Rate (SR), Success-weighted Path Length (SPL), and Success-weighted Action Efficiency (SAE), reported in Table~\ref{tab:realrobot-agg}.

\begin{table}[h]
\centering
\footnotesize
\caption{GRIP-R Real-World Performance}
\begin{tabular}{l|c|c|c}
\hline
\textbf{Target Object} & \textbf{SR (\%)} & \textbf{SPL (\%)} & \textbf{SAE (\%)} \\
\hline
Laptop         & 86.67 & 91.00 & 82.00 \\
TV             & 93.33 & 90.00 & 91.00 \\
Refrigerator   & 100.00 & 97.00 & 98.00 \\
Sports Ball    & 93.33 & 93.39 & 97.61 \\
Backpack       & 93.33 & 90.00 & 84.00 \\
\hline
\end{tabular}
\label{tab:realrobot-agg}
\end{table}

\begin{figure*}[t]
  \centering
  \subfloat[Laptop: Anchored via \textit{Chair}, \textit{Keyboard}.]%
  {\includegraphics[width=0.19\textwidth]{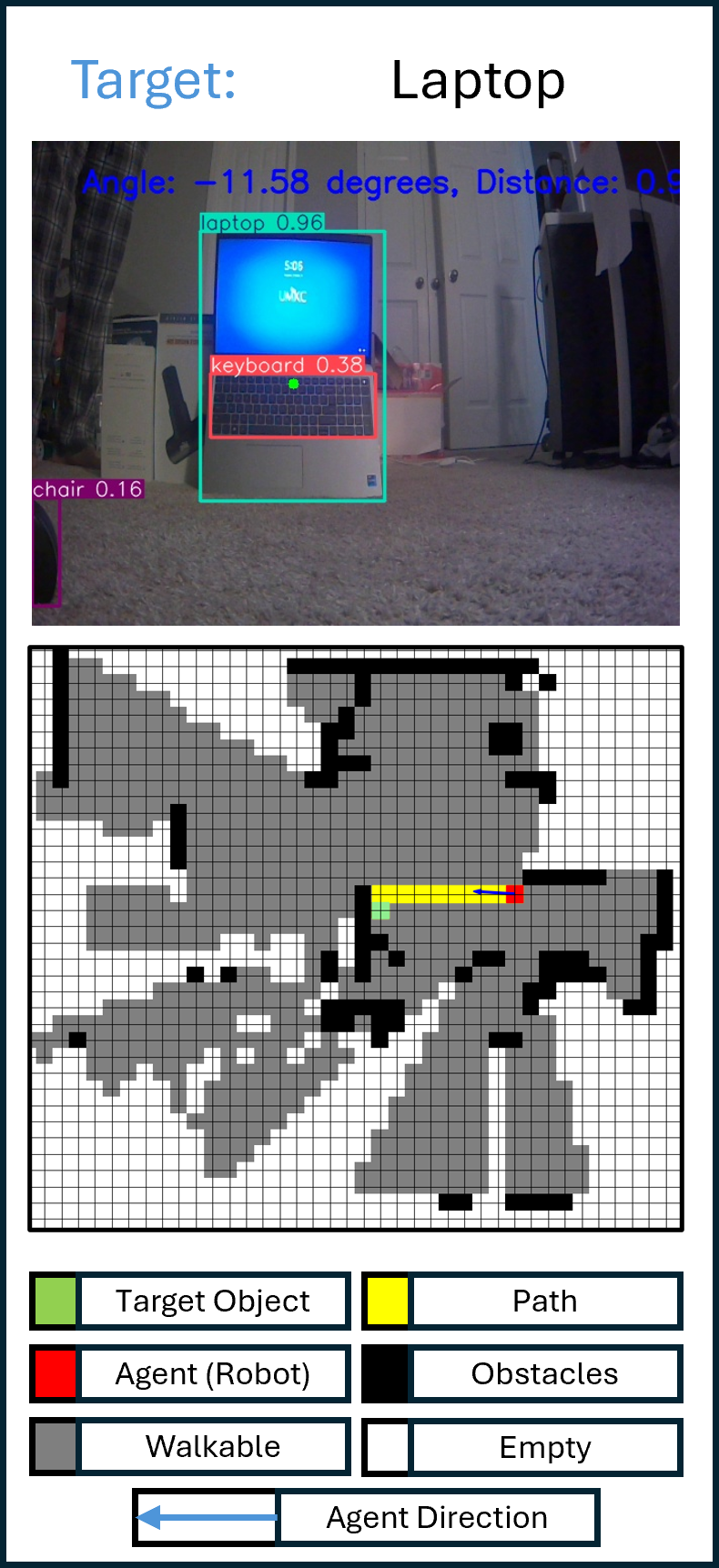}\label{fig:realrobot-a}}
  \hfil
  \subfloat[TV: Detected at $-7.82^\circ$.]%
  {\includegraphics[width=0.19\textwidth]{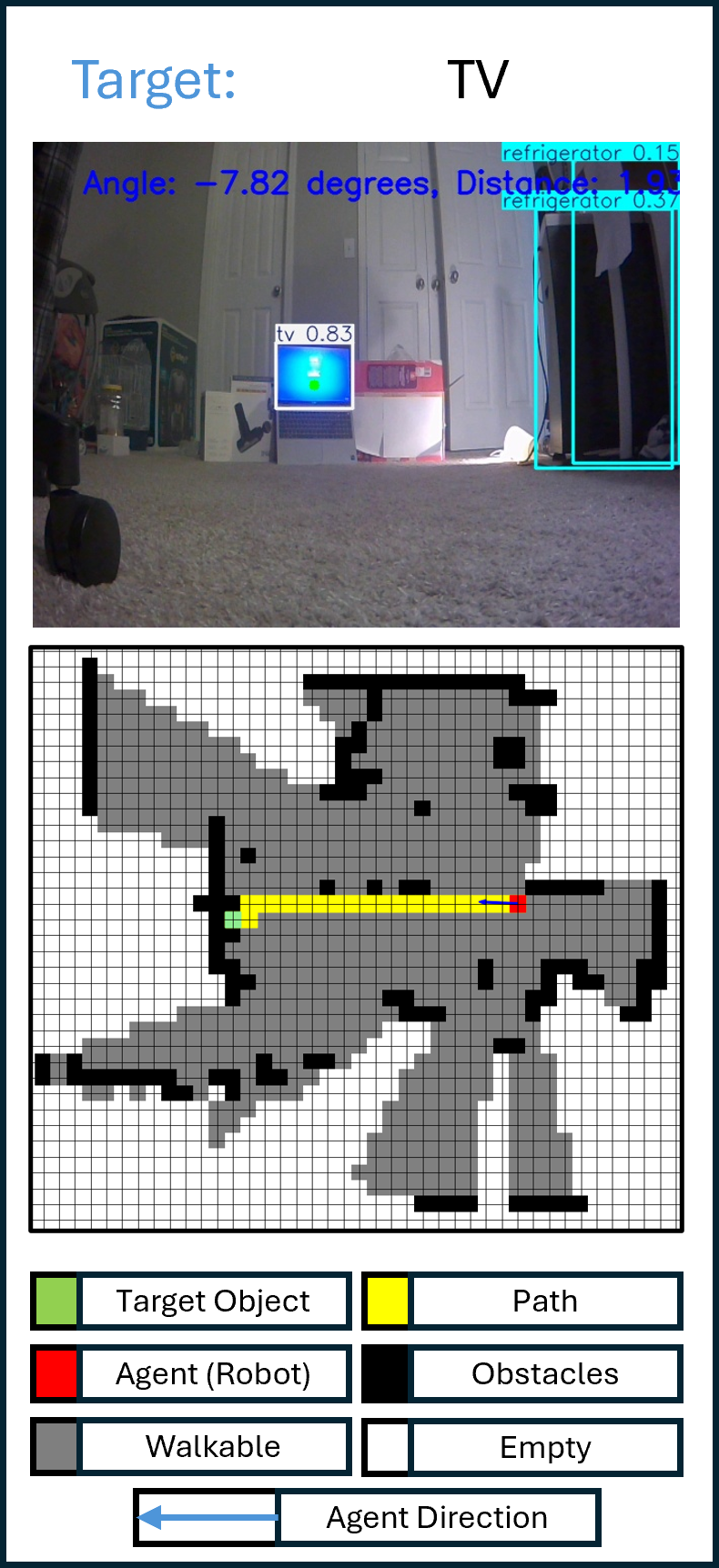}\label{fig:realrobot-b}}
  \hfil
  \subfloat[Refrigerator: Efficient, direct path.]%
  {\includegraphics[width=0.19\textwidth]{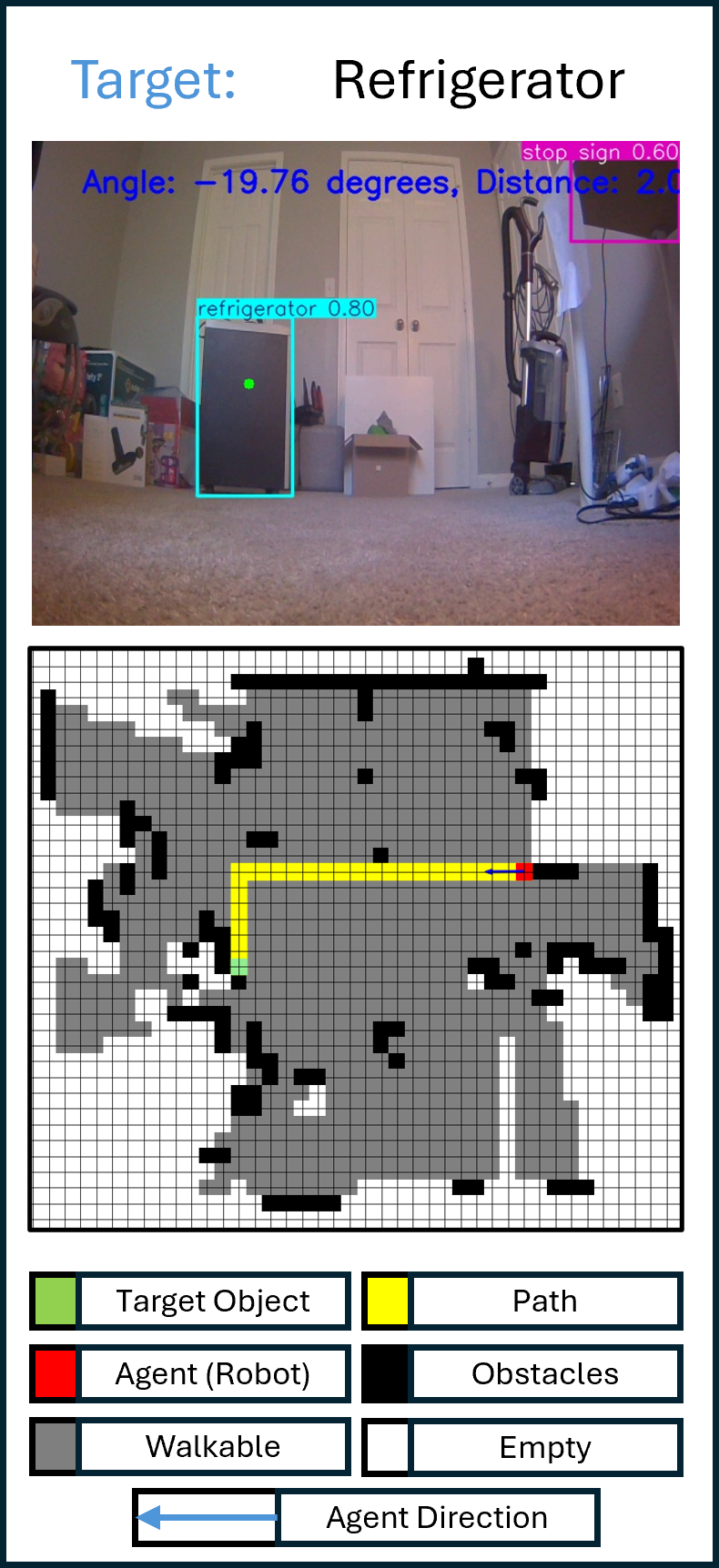}\label{fig:realrobot-c}}
  \hfil
  \subfloat[Sports Ball: Reached despite occlusion.]%
  {\includegraphics[width=0.19\textwidth]{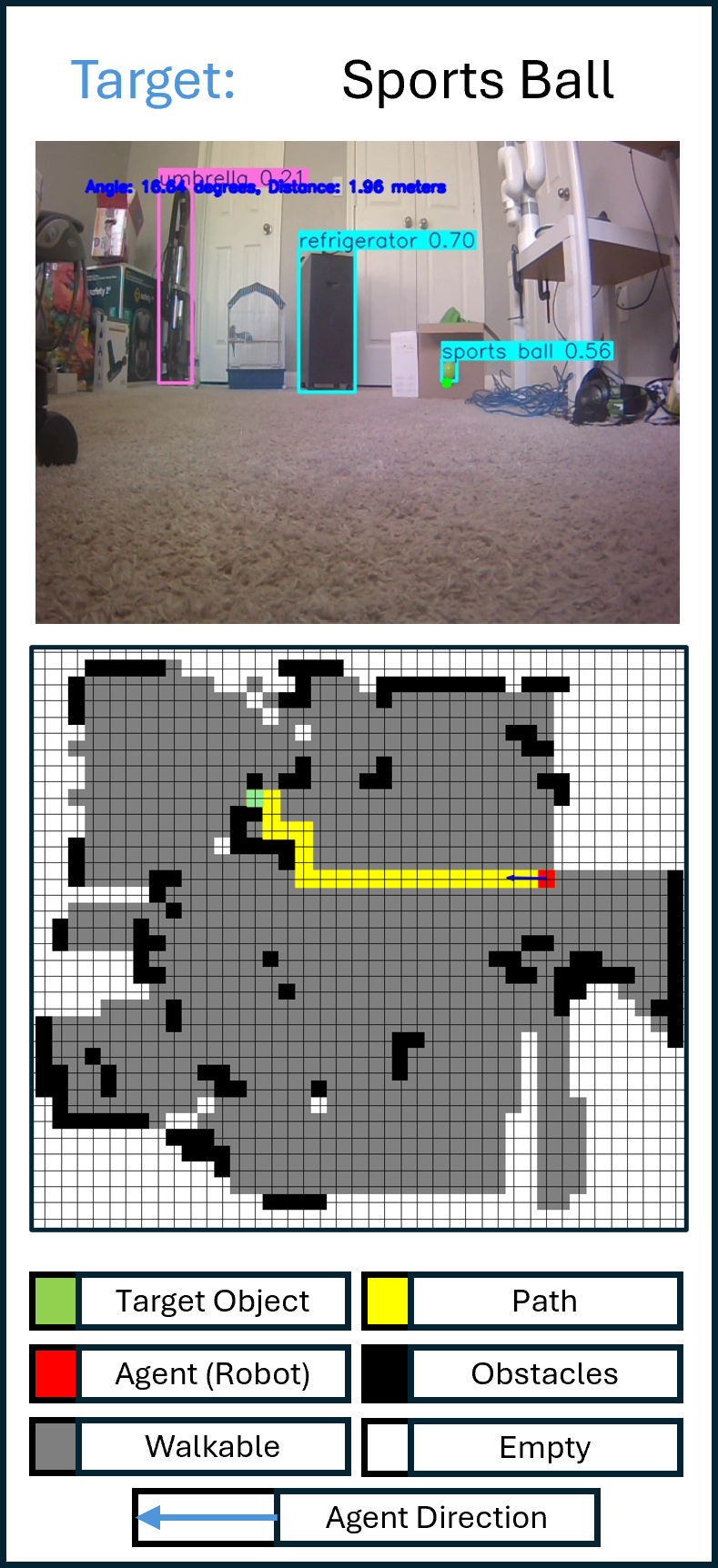}\label{fig:realrobot-d}}
  \hfil
  \subfloat[Backpack: Located via \textit{Suitcase} anchor.]%
  {\includegraphics[width=0.19\textwidth]{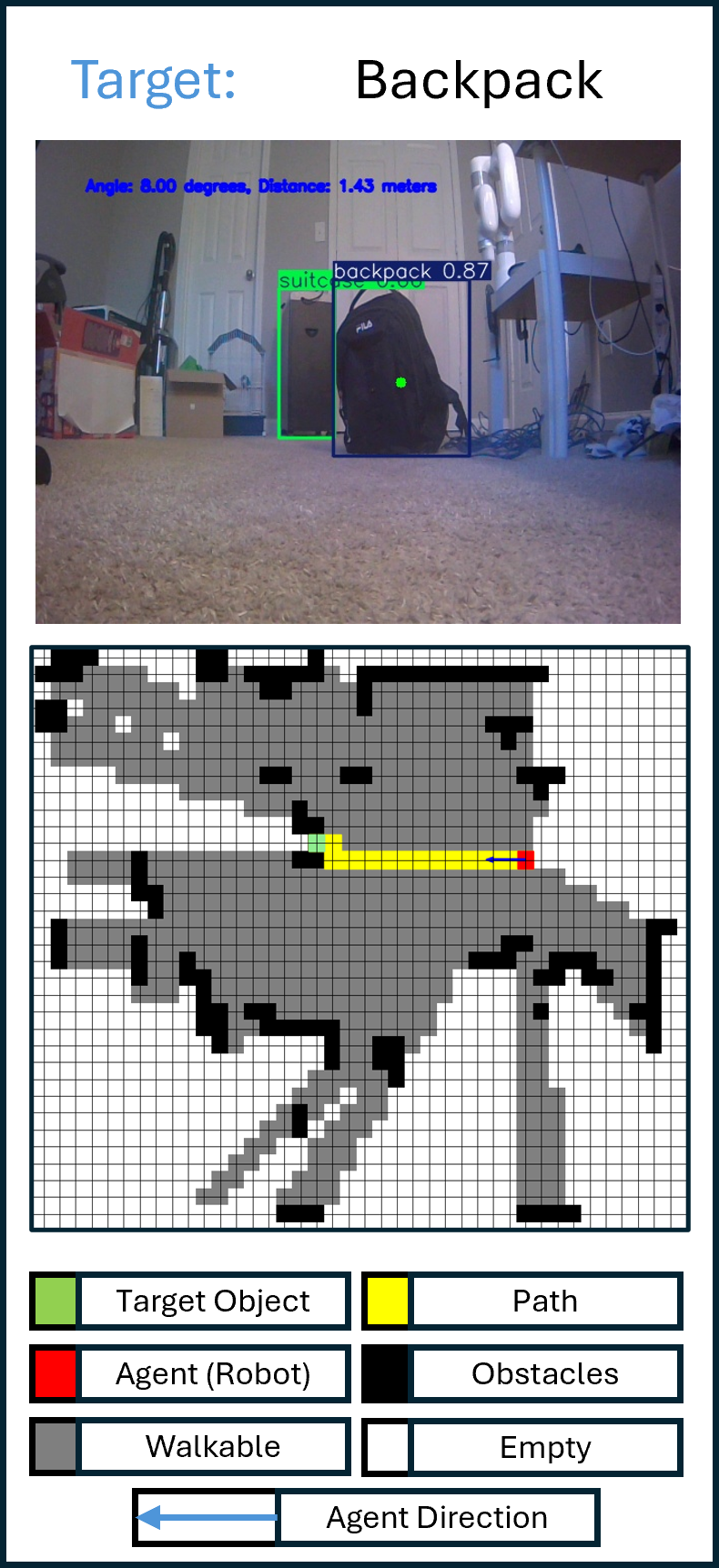}\label{fig:realrobot-e}}
  \caption{GRIP-R real-world execution. Symbolic reasoning enables robust navigation under uncertainty.}
  \label{fig:realrobot}
\end{figure*}

\noindent
Figure~\ref{fig:realrobot} shows examples of symbolic navigation. Anchors like \textit{Chair}, \textit{Keyboard}, and \textit{Suitcase} were auto-selected via the co-occurrence graph to guide under partial observability.



\section{Failure Case Analysis and Symbolic Introspection in GRIP-F}
\label{sec:failure-analysis}

To investigate the limits of GRIP-F’s symbolic planning, we conducted a detailed failure analysis on RoboTHOR episodes where the agent was unable to complete the object-goal navigation task despite a valid symbolic plan. These \textit{hard failure} cases reveal bottlenecks in spatial feasibility, anchor visibility, and symbolic-to-spatial alignment.

\subsection{Common Failure Patterns}

Our curated failure set exposes three recurring failure types:

\begin{itemize}
    \item \textit{Unreachable Anchors:} Co-occurrence-based symbolic anchors such as \textit{Drawer} are semantically plausible but often inaccessible due to environment layout constraints.
    
    \item \textit{Visual Misprioritization:} Symbolic reasoning may overweight semantically relevant but visually absent anchors, leading to inefficient trajectories or drift near the goal.
    
    \item \textit{Layout Mismatches:} The symbolic path may assume spatial configurations (e.g., adjacency between objects) that do not hold in the actual scene.
\end{itemize}

\subsection{Case Study Examples}

\begin{figure*}[ht!]
    \centering
    \includegraphics[width=0.9\textwidth]{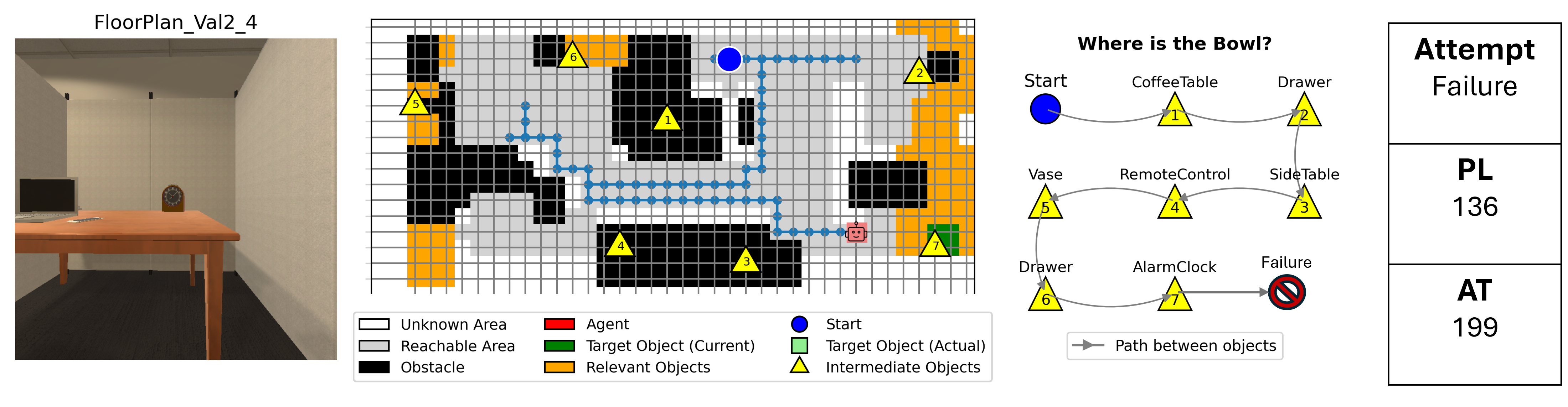}
    \vspace{0.5em}
    
    \includegraphics[width=0.9\textwidth]{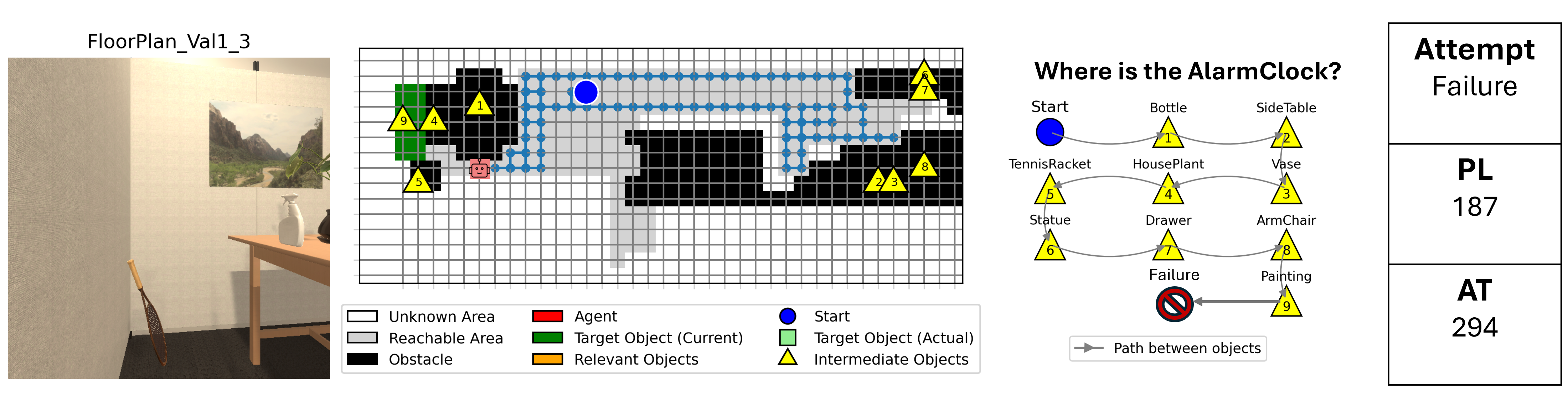}
    \vspace{0.5em}
    
    \includegraphics[width=0.9\textwidth]{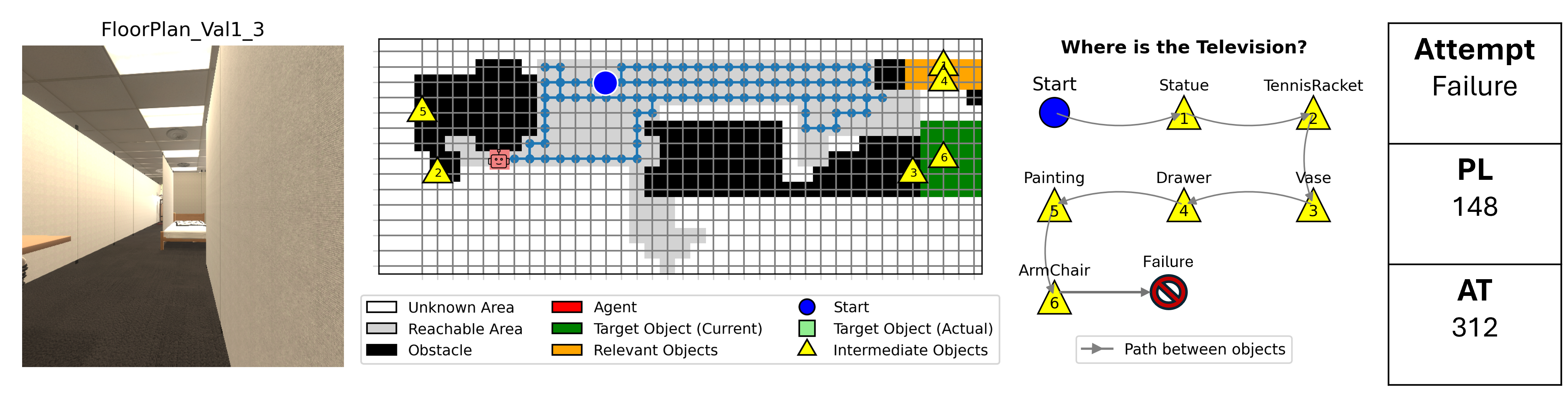}
    \caption{Failure Cases: GRIP-F fails due to semantic-spatial mismatches. Anchors were semantically relevant but spatially unreachable or misleading.}
    \label{fig:failure_cases}
\end{figure*}

Figure~\ref{fig:failure_cases} illustrates three GRIP-F failure scenarios. In each case, symbolic chaining proceeds with interpretable anchors, but spatial execution fails due to layout or visibility issues.

\subsection{Symbolic Chain Introspection}

GRIP-F produces interpretable symbolic sequences—e.g., \textit{Vase} $\rightarrow$ \textit{Chair} $\rightarrow$ \textit{Desk}—which function both as plans and introspective diagnostics. Table~\ref{tab:failure_case_metrics} summarizes these cases, linking symbolic reasoning with performance breakdowns.

\begin{table*}[ht!]
\centering
\footnotesize
\caption{Failure Case Summary with Symbolic Planning Chains}
\label{tab:failure_case_metrics}
\begin{tabular}{@{}p{1cm} l c c c c c l p{7.8cm}@{}}
\toprule
\textbf{Case} & \textbf{Target} & \textbf{Result} & \textbf{Path Len} & \textbf{Act.} & \textbf{Type} & \textbf{Symbolic Reasoning Summary} \\
\midrule
Val2\_4 & Bowl       & Failure & 136 & 199 & Hard & Anchor \textit{Drawer} was unreachable despite valid semantic relation. \\
Val1\_3 & Alarm Clock& Failure & 187 & 294 & Hard & Missed nearby target due to drift toward non-visible anchors. \\
Val1\_3 & Television & Failure & 148 & 312 & Hard & Layout mismatch caused agent to bypass the actual target. \\
\bottomrule
\end{tabular}
\end{table*}

\subsection{Limitations and Future Enhancements}

These findings emphasize the need for:

\begin{itemize}
    \item \textit{Visibility-aware anchor filtering}, to eliminate semantically plausible but visually blocked anchors.
    \item \textit{Confidence-based anchor scoring}, integrating spatial feasibility into symbolic planning.
    \item \textit{LLM-guided symbolic repair}, enabling dynamic replanning when anchor assumptions break down.
\end{itemize}

This failure-driven introspection highlights a key design insight: while GRIP-F’s symbolic chaining is interpretable and effective in many settings, its static planning can benefit from adaptive mechanisms for robustness in real-world layouts.

\section{Discussion}
GRIP provides a unified framework that integrates spatial, symbolic, and linguistic reasoning for object-goal navigation, with three variants—GRIP-L, GRIP-F, and GRIP-R—tailored for increasing environmental complexity.

\textit{GRIP-L} is optimized for structured, perception-driven settings like AI2-THOR. It uses co-occurrence-based anchor selection and fine-grained grid planning to outperform perception-only baselines at low computational cost. However, it lacks multi-hop planning and introspective recovery, limiting performance under occlusion or ambiguity.

\textit{GRIP-F} adds symbolic relay chaining for mid-horizon planning in cluttered environments like RoboTHOR. It composes interpretable anchor sequences (e.g., \textit{Vase} $\rightarrow$ \textit{Chair} $\rightarrow$ \textit{Desk}) to guide navigation under partial observability. Still, it lacks visibility filtering and anchor confidence scoring, leading to occasional symbolic overhops and execution failures.

\textit{GRIP-R} validates GRIP’s symbolic reasoning in real-world deployment using a Jetbot Pro in multi-room apartments. It performs well despite sensor noise and lighting variations, though it currently depends on static semantic priors without LLM-based replanning or depth-aware filtering.

\textit{Comparison with Prior Work.} GRIP outperforms traditional ObjectNav approaches that use coarse spatial grids~\cite{chaplot2020object, du2020learning}, rigid priors~\cite{maksymets2021thda}, or opaque reasoning~\cite{khanna2024goat, xu2024aligning}. Unlike models like SEAL~\cite{chaplot2021seal} or CL-CoTNav~\cite{cai2025cl}, GRIP supports online, interpretable planning using a 5\,cm-resolution grid, a sparse symbolic graph, and GPT-4o-powered introspection.

\vspace{0.5em}
\noindent
\textit{Limitations and Future Work.}
\begin{itemize}
    \item \textit{GRIP-L:} Incorporate lightweight chaining and introspection for improved range and adaptability.
    \item \textit{GRIP-F:} Add visibility-aware anchor filtering, confidence scoring, and dynamic replanning via LLMs.
    \item \textit{GRIP-R:} Extends to varied environments with RGB-D sensing, anchor validation, and GPT-based recovery.

\end{itemize}

GRIP’s modular integration of perception, symbolic planning, and language enables interpretable, transferable navigation strategies. It balances performance with explainability, key for robust embodied agents in dynamic settings.

\section{Conclusion}

We present \textit{GRIP}, a unified and interpretable framework for object-goal navigation that combines semantic grounding, symbolic planning, and real-world execution. GRIP uniquely supports online construction of instruction-aligned semantic graphs, closed-loop D$^*$-based replanning, symbolic chaining, and LLM-guided introspection—capabilities previously scattered across disparate systems.

GRIP is instantiated in three modular variants. The first, \textit{GRIP-L (Lightweight)}, enables symbolic planning over static semantic graphs in simulation, offering a streamlined baseline for structured environments. The second, \textit{GRIP-F (Full)}, introduces dynamic subgoal chaining, introspective reasoning, and execution-aware replanning, allowing it to interpret ambiguous instructions and recover from failure. The third, \textit{GRIP-R (Real-World)}, is deployed on a Jetbot Pro robot equipped with RGB, LiDAR, and IMU sensors, enabling real-world navigation supported by GPT-4o–guided symbolic adaptation and plan repair.

Through extensive evaluations on the AI2-THOR and RoboTHOR benchmarks, GRIP demonstrates robust generalization across diverse environments, task complexities, and failure conditions. Unlike prior systems, GRIP is the only framework—according to Table~\ref{tab:unified_objectnav_comparison}—that simultaneously supports fine-grained semantic grid construction, relational symbolic graph generation, dynamic replanning through D$^*$ under execution failures, integration of open-vocabulary reasoning via large language models, and full deployment on physical robots navigating cluttered indoor scenes.

Despite its strengths, GRIP has several limitations. It currently lacks visibility-aware filtering to prioritize anchors based on observability, dynamic confidence calibration for evaluating symbolic subgoal reliability, and proactive introspection for speculative plan refinement. Additionally, GRIP-R’s deployment remains constrained to a single real-world testbed and a limited object vocabulary. Future work will address these gaps by integrating depth-informed planning heuristics, expanding deployment across diverse environments, and enabling more expressive, conversational planning via LLMs.
By bridging symbolic abstraction, multimodal perception, and embodied reasoning in a single architecture, GRIP sets a new benchmark for interpretable, adaptable, and physically grounded object-goal navigation.


\balance
\bibliographystyle{unsrt}
\bibliography{New-References}

\newpage

\appendix
\section{Object-Wise Evaluation Across Benchmarks}
\label{app:objectwise-eval}

To evaluate GRIP’s robustness and object-level generalization, we conduct a detailed comparison across target object categories using the three GRIP variants:  
\begin{itemize}
    \item \textit{GRIP-L}, deployed in AI2-THOR with lightweight symbolic planning,
    \item \textit{GRIP-F}, used in RoboTHOR with full symbolic chaining and semantic relay integration, and
    \item \textit{GRIP-R}, deployed in real-world environments.
\end{itemize}

\begin{figure*}[hb]
  \centering
  \includegraphics[width=0.9\textwidth]{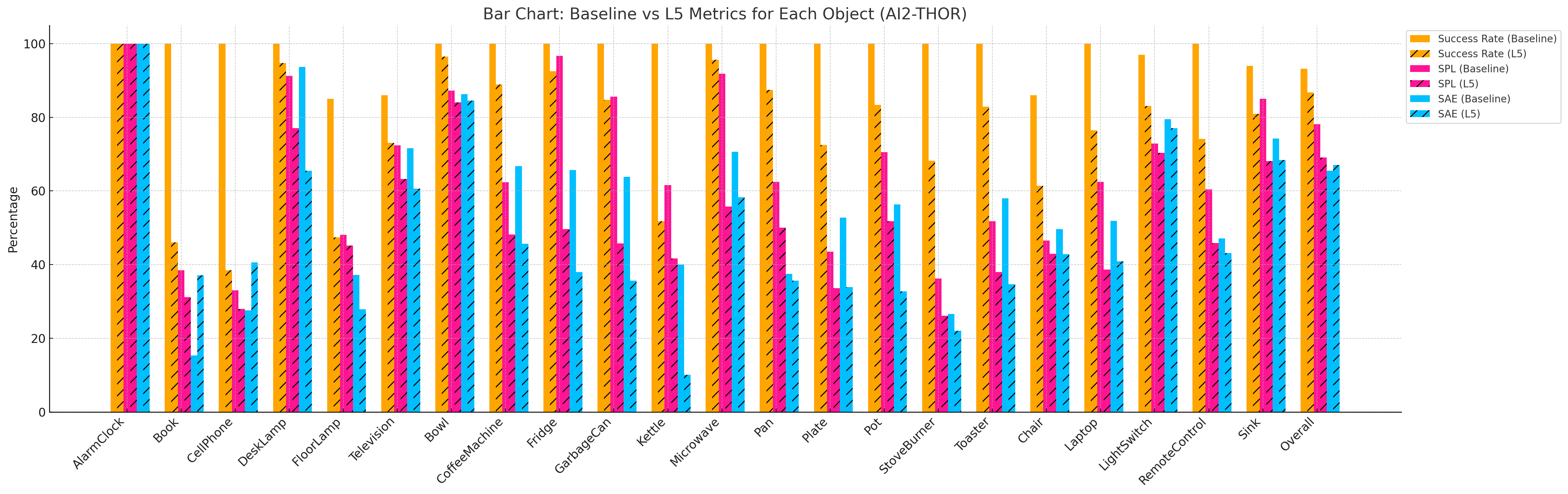}
  \vspace{1em}
  \includegraphics[width=0.85\textwidth]{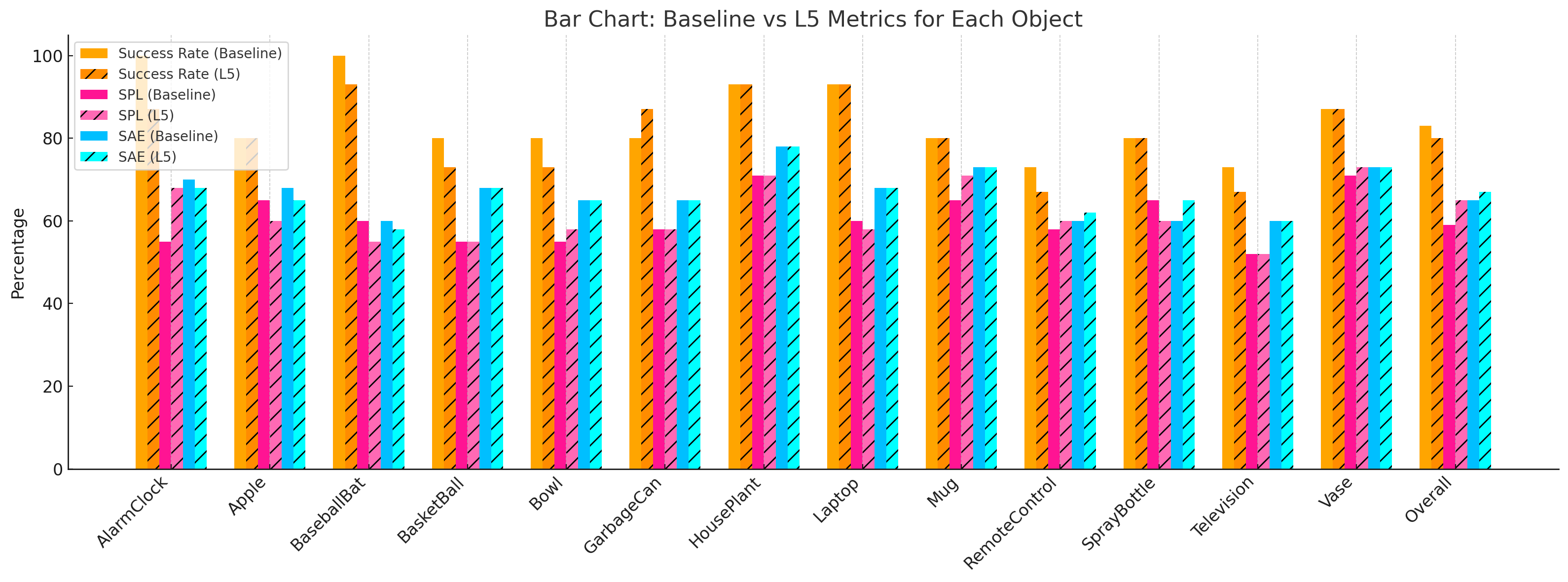}
  \caption{\textit{Object-Wise Comparison of SR, SPL, and SAE.} \textit{Top:} GRIP-L performance in AI2-THOR. \textit{Bottom:} GRIP-F performance in RoboTHOR. Metrics are reported per object class for baseline and L5-enhanced configurations.}
  \label{fig:objectwise-barplots}
\end{figure*}

This section focuses on simulation-based evaluation. We report \textit{Success Rate (SR)}, \textit{Success-weighted Path Length (SPL)}, and \textit{Success-weighted Action Efficiency (SAE)} across 21 object classes in AI2-THOR and 13 in RoboTHOR. Object-level performance for GRIP-L and GRIP-F is illustrated in Figure~\ref{fig:objectwise-barplots}.

\paragraph{AI2-THOR (GRIP-L): Favorable to Perception-Dominant Models}  
GRIP-L, operating in AI2-THOR, performs well across objects with high visibility, such as \textit{AlarmClock}, \textit{Sink}, and \textit{DeskLamp}, where simple visual grounding suffices. However, SPL and SAE decline for occluded or spatially ambiguous objects like \textit{Toaster} and \textit{Chair}, where deeper symbolic reasoning would be beneficial. The L5-enhanced GRIP-L variant introduces modest gains through improved semantic filtering and trajectory pruning, though the environment’s clarity limits the impact of symbolic relays.

\paragraph{RoboTHOR (GRIP-F): Emphasizing Symbolic and Spatial Reasoning}  
GRIP-F’s evaluation in RoboTHOR reflects stronger reliance on mid-horizon semantic inference. The environment’s occlusions and complex layouts highlight GRIP-F’s strengths: symbolic relay chaining, co-occurrence-based subgoal prediction, and context-aware navigation. Notable improvements in SR, SPL, and SAE are observed for objects like \textit{Laptop}, \textit{Mug}, and \textit{HousePlant}, which often require disambiguation or intermediate anchoring to reach.

\begin{itemize}
  \item \textit{Environment-Specific Behavior:} GRIP-L excels in AI2-THOR with clear visibility, while GRIP-F shows superior planning in RoboTHOR’s occluded, multi-room scenes.

  \item \textit{Semantic Relay Impact:} GRIP-F’s symbolic chaining yields SPL and SAE improvements in RoboTHOR, especially for long-horizon and ambiguous goals.

    \item \textit{Adaptive Reasoning Strategies:} GRIP dynamically adjusts its depth of symbolic abstraction based on environment demands—leveraging direct perception in GRIP-L and strategic inference in GRIP-F.
    
    \item \textit{Module Extensibility:} Though not evaluated here, GRIP-R applies similar symbolic grounding and co-occurrence anchoring in the real world, with LLM-based recovery held for separate analysis (see Section~\ref{subsec:realworld-eval}).
\end{itemize}

\vspace{0.5em}
\noindent
These findings highlight the complementary roles of GRIP-L and GRIP-F across structured and complex benchmarks, reinforcing GRIP’s modular design as scalable from perception grounding to symbolic reasoning for robust object-goal navigation across diverse environments.

\section*{Appendix A: Successful Symbolic Chains in GRIP-F}
\label{appendix:success-cases}

To complement our failure analysis, we present representative \textit{hard success} cases where GRIP-F successfully navigated to occluded or partially visible targets using symbolic chaining. These examples highlight the model’s capacity for multi-hop semantic reasoning under limited visual input.

\begin{figure*}[ht!]
    \centering
    \includegraphics[width=0.85\textwidth]{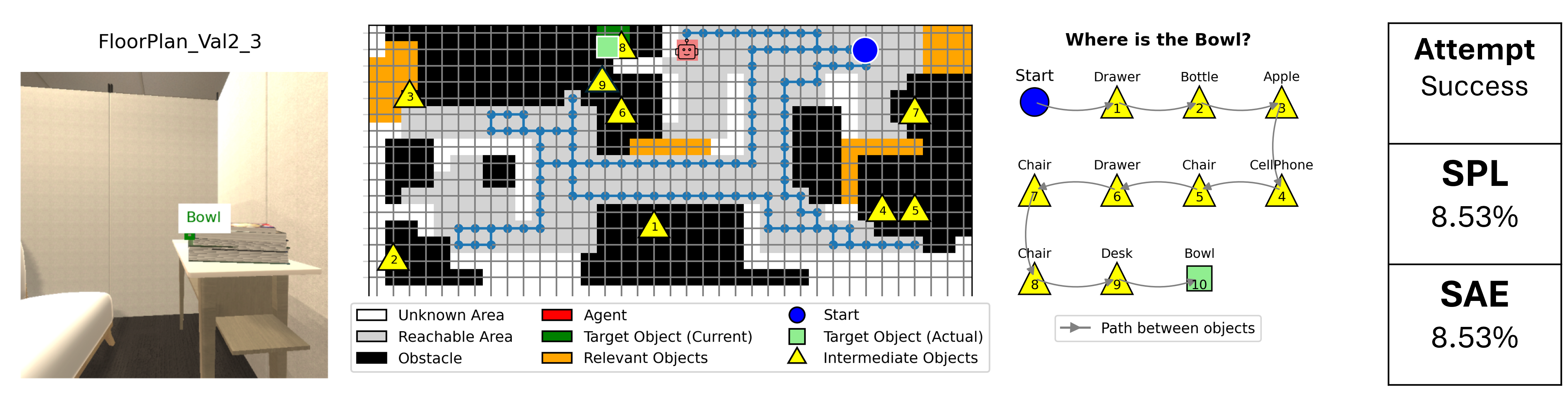}
    \vspace{0.5em}
    
    \includegraphics[width=0.85\textwidth]{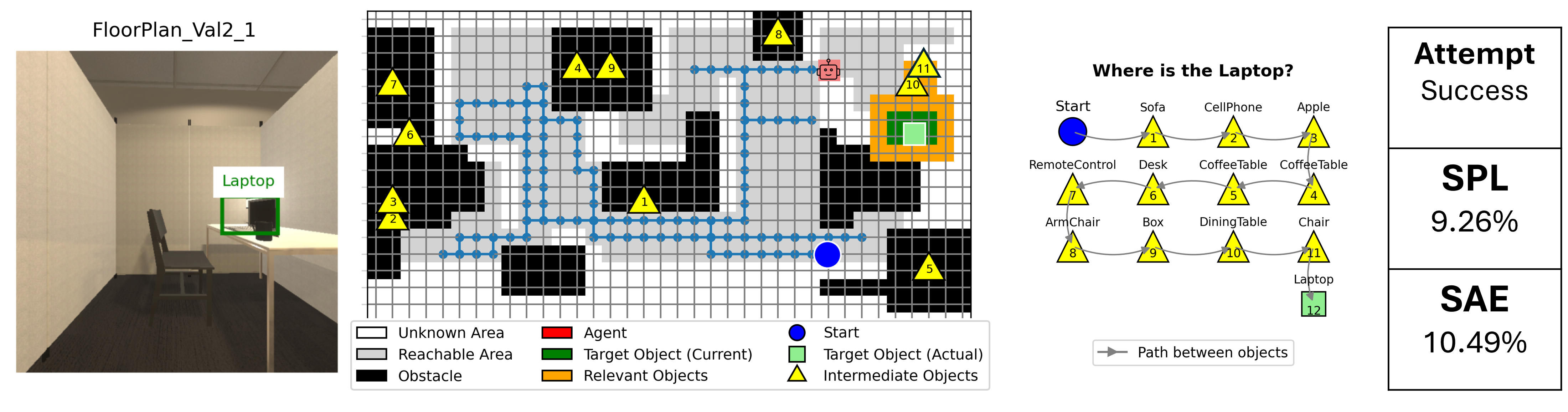}
    \vspace{0.5em}
    
    \includegraphics[width=0.85\textwidth]{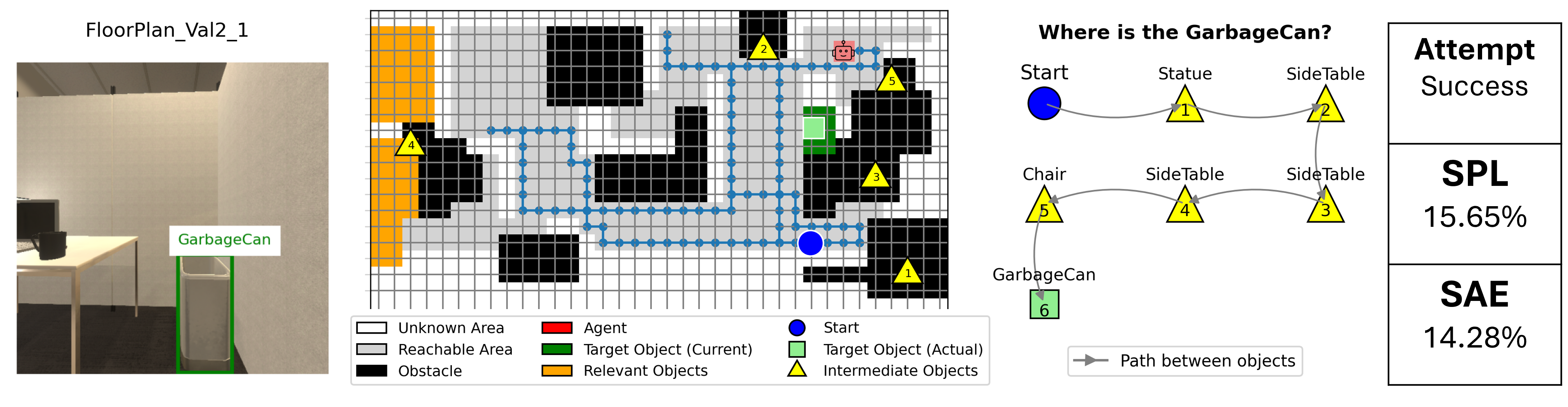}
    \caption{Success Cases: GRIP-F reaches targets using symbolic anchor sequences despite visual ambiguity.}
    \label{fig:success_cases}
\end{figure*}

Figure~\ref{fig:success_cases} shows three cases in which GRIP-F used symbolic anchors to overcome occlusion and ambiguity.

\begin{table*}[!t]
\centering
\footnotesize
\caption{Successful GRIP-F Episodes: Symbolic Planning Chains and Metrics}
\label{tab:success_case_metrics}
\begin{tabular}{@{}p{0.8cm} l c c c c c l p{7.8cm}@{}}
\toprule
\textbf{Case} & \textbf{Target} & \textbf{Result} & \textbf{SPL} & \textbf{SAE} & \textbf{Path Len} & \textbf{Act.} & \textbf{Type} & \textbf{Symbolic Reasoning Summary} \\
\midrule
Val2\_3 & Bowl       & Success & 8.53\%  & 8.53\%  & 53  & 71  & Hard & Used symbolic chain \textit{Vase} $\rightarrow$ \textit{Chair} $\rightarrow$ \textit{Desk} to bypass occlusion. \\
Val2\_1 & Laptop     & Success & 9.26\%  & 10.49\% & 61  & 78  & Hard & Navigated ambiguous space using mid-horizon anchor selection. \\
Val2\_1 & Garbagecan & Success & 15.65\% & 14.28\% & 67  & 82  & Hard & Anchored to \textit{Side Table} for disambiguation in cluttered layout. \\
\bottomrule
\end{tabular}
\end{table*}

\begin{itemize}
    \item \textit{Val2\_3 (Bowl):} Target not initially visible. GRIP-F inferred a symbolic path \textit{Vase} $\rightarrow$ \textit{Chair} $\rightarrow$ \textit{Desk}, successfully reaching the occluded object.
    
    \item \textit{Val2\_1 (Laptop):} GRIP-F leveraged partial anchors and semantic priors to navigate through a cluttered scene.

    \item \textit{Val2\_1 (Garbagecan):} Selected \textit{Side Table} as a mid-horizon anchor for spatial disambiguation.
\end{itemize}


Table~\ref{tab:success_case_metrics} summarizes navigation performance and symbolic reasoning for each successful case.

\end{document}